\documentclass{article}

\PassOptionsToPackage{table}{xcolor}

\usepackage[preprint]{neurips_2026}

\usepackage{tikz}
\usepackage{fontawesome5}

\usepackage{amsmath,amsfonts,bm}

\def\eqref#1{equation~\ref{#1}}

\def\1{\bm{1}}

\DeclareMathAlphabet{\mathsfit}{\encodingdefault}{\sfdefault}{m}{sl}
\SetMathAlphabet{\mathsfit}{bold}{\encodingdefault}{\sfdefault}{bx}{n}

\usepackage[hidelinks]{hyperref}

\usepackage{latexsym}

\usepackage[T1]{fontenc}
\usepackage[utf8]{inputenc}

\usepackage{microtype}

\usepackage{inconsolata}

\usepackage{xcolor}

\usepackage{graphicx}

\usepackage{wrapfig}

\usepackage{amsmath}
\usepackage{amssymb}

\usepackage{algorithm}
\usepackage{algpseudocode}

\usepackage{booktabs} 
\usepackage{array}    
\usepackage{subcaption} 

\usepackage{enumitem}

\definecolor{topcolor}{HTML}{FFF2CC}
\definecolor{secondcolor}{HTML}{CFE2F3}

\definecolor{oursrow}{HTML}{DDEBF7}
\definecolor{deltarow}{HTML}{E9E9F6}

\definecolor{algcomment}{HTML}{2E6DB4}

\usepackage{xspace}

\newcommand{\methodexp}{\textbf{A}daptive \textbf{H}indsight with \textbf{E}nvironment-\textbf{A}ugmented \textbf{D}istillation}

\usepackage[most]{tcolorbox}
\definecolor{promptbar}{HTML}{6B6B6B}
\newtcolorbox{promptbox}[1]{
  enhanced,
  colback=white, colframe=promptbar,
  coltitle=white, colbacktitle=promptbar,
  fonttitle=\bfseries\small, fontupper=\small,
  title={#1},
  boxrule=0.9pt, arc=0pt, outer arc=0pt,
  left=7pt, right=7pt, top=6pt, bottom=6pt,
}
\newcommand{\slot}[1]{\{\texttt{#1}\}}
\newcommand{\xtag}[1]{\texttt{\textless#1\textgreater}}

\definecolor{aheadteal}{RGB}{92, 52, 170}

\newcommand{\aheadicon}{%
  \raisebox{-0.18\height}{%
    \includegraphics[height=1.8em]{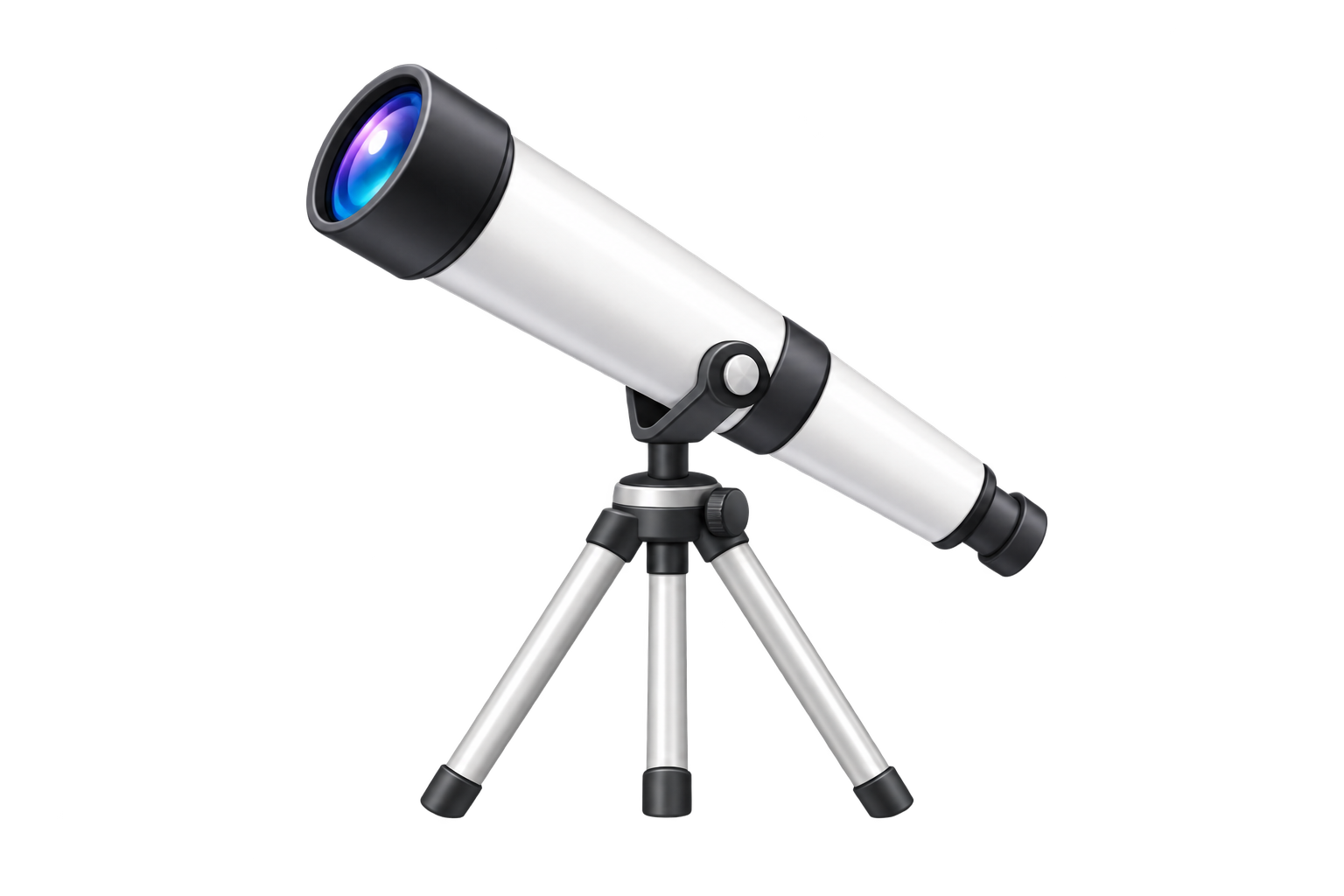}%
  }%
}

\newcommand{\method}{%
  \textcolor{aheadteal}{\textsc{Ahead}}\xspace%
}

\newcommand{\methodtitle}{%
  \textcolor{aheadteal}{A}daptive
  \textcolor{aheadteal}{H}indsight with
  \textcolor{aheadteal}{E}nvironment-\textcolor{aheadteal}{A}ugmented
  \textcolor{aheadteal}{D}istillation\xspace%
}

\title{
  \aheadicon\;
  \method: \methodtitle
  for Agentic RL
}

\author{
Xiaolong Jin$^{1,2}$\thanks{Work done during an internship at Amazon.},~~~Dingmin Wang$^{1}$\thanks{Corresponding author.}, ~~~Vijay Lingam$^{1}$, ~Varun Kumar$^{1}$ \\
$^{1}$AWS AI Labs \quad $^{2}$Purdue University
}

\definecolor{buttonborder}{RGB}{190,205,215}
\definecolor{projectblue}{RGB}{30,100,230}
\definecolor{modelgray}{RGB}{80,95,115}

\newcommand{\linkbutton}[3]{%
  \href{#1}{%
    \tikz[baseline=-0.6ex]{
      \node[
        draw=buttonborder,
        rounded corners=8pt,
        line width=0.5pt,
        inner xsep=8pt,
        inner ysep=3pt,
        font=\bfseries
      ] {#2\hspace{3pt} #3};
    }%
  }%
}

\newcommand{\hficon}{%
  \raisebox{-0.2\height}{%
    \includegraphics[height=1.05em]{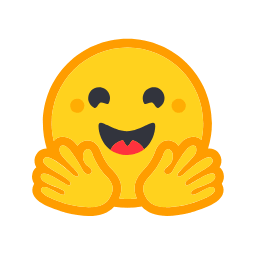}%
  }%
}

\begin{document}
\maketitle

\vspace{-1.0em}

\begin{center}
\linkbutton{https://jinxiaolong1129.github.io/AHEAD/}
  {\textcolor{projectblue}{\faGlobe}}
  {Project Page}
\hspace{3em}
\linkbutton{https://huggingface.co/collections/Bruce-Jin/ahead-alfworld-and-webshop-agents}
  {\hficon}
  {Model Card}
\end{center}

\vspace{0.5em}

\begin{abstract}
Training multi-turn LLM agents with reinforcement learning typically relies on trajectory-level rewards, which assign a uniform advantage to every step and cannot identify which decisions led to success or failure. Self-distillation methods can provide finer-grained supervision by augmenting RL with privileged information. However, existing approaches usually apply the same type of privileged information to every step in an indistinguishable manner, ignoring a key asymmetry: routine steps need little additional guidance, while critical error steps require corrective direction that environment feedback alone cannot provide. We propose \method, a step-aware framework that matches different supervision sources to different step types. The teacher receives environment feedback on all steps as a grounded dense signal, and additionally receives LLM-generated corrective hints on error steps to supply the direction that environment feedback lacks. The method introduces minimal changes to the standard GRPO algorithm. Across ALFWorld, WebShop, and Search-based QA, and across three model scales, \method raises task success ($+13.3$ points on ALFWorld and $+11.0$ on WebShop at 7B over GRPO), reaches a given success rate in fewer training steps, and solves tasks within tighter interaction budgets than outcome-only RL and prior self-distillation baselines.

\end{abstract}

\section{Introduction}
Reinforcement learning has become a central approach for post-training multi-turn LLM agents~\citep{grpo,dong2025arpo,feng2025gigpo}.
Unlike in single-turn reasoning, multi-turn agents interact with environments over extended horizons, where each action changes future observations and shapes subsequent decisions.
Methods such as GRPO~\citep{grpo} train policies from trajectory-level environment rewards without requiring a critic, but assign a uniform advantage to every token in the trajectory.
However, not all steps contribute equally to the outcome.
Most are routine: \textit{the agent acts reasonably and the environment progresses normally}.
A small number of steps, however, largely determine the outcome; while a wrong object selection, an invalid action, or a misguided search query can compromise the entire episode.
Assigning the same gradient signal to both types provides no mechanism to correct the decisions that actually matter.

On-policy self-distillation offers denser supervision by augmenting a teacher branch with privileged information (PI) and comparing its token-level predictions against the unaugmented student~\citep{opsd,rlsd,sdar}, producing a log-probability gap that serves as a token-level credit-assignment signal.
The informativeness of this signal, however, depends entirely on what PI the teacher receives.
Existing methods use task-level PI, such as retrieved skills~\citep{sdar,wang2026skillsd} or reference answers~\citep{opsd}, that provides the same guidance to every step in the trajectory.
This ignores a key asymmetry: routine steps need only confirmation that the action was appropriate, while error steps need specific corrective direction telling the agent what it should have done instead.
Task-level PI, which summarizes general workflows rather than diagnosing individual decisions, cannot provide such step-specific corrective information.

\begin{figure}[t]
  \centering
  \includegraphics[width=0.95\textwidth]{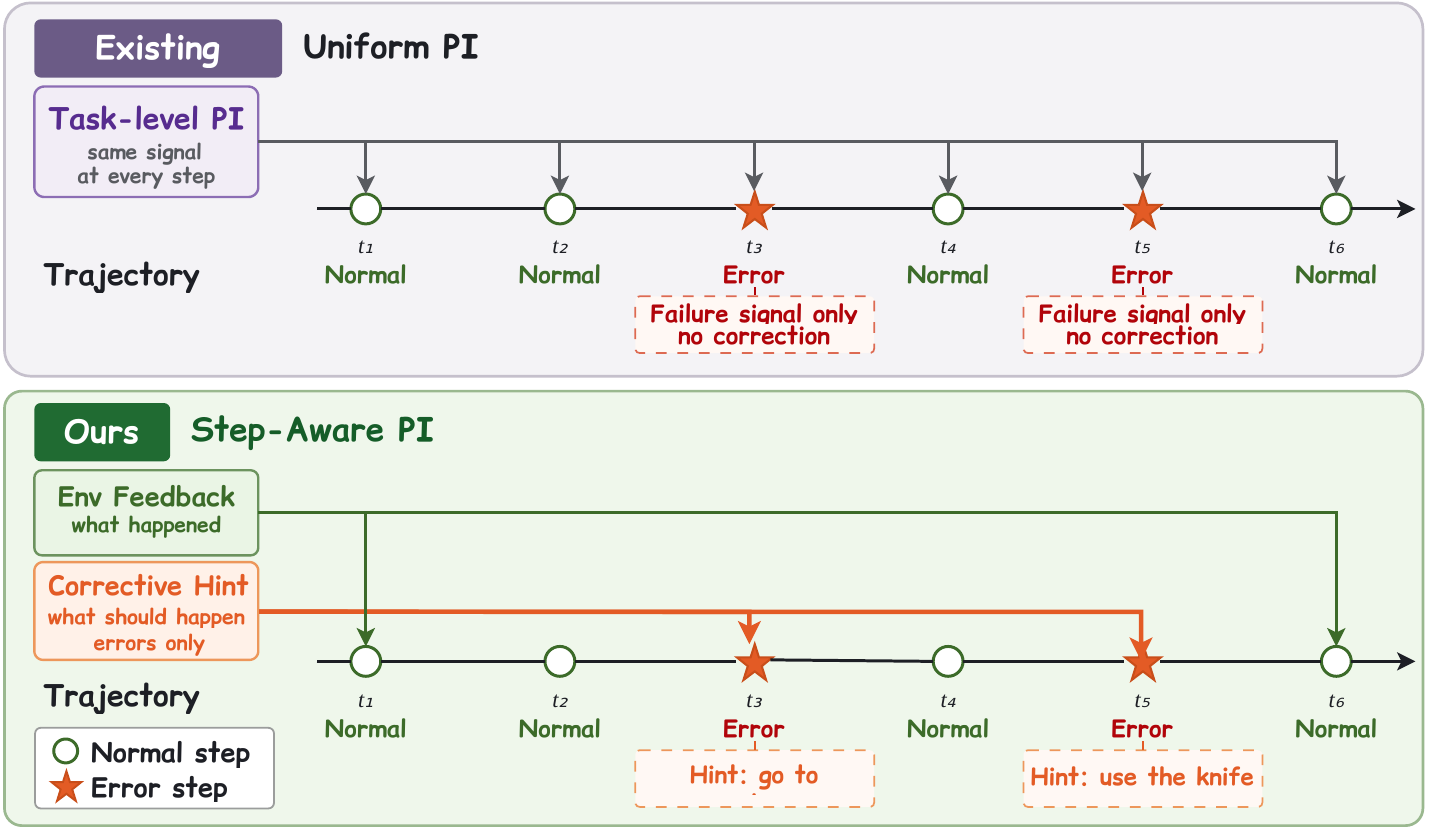}
  \caption{Existing self-distillation methods apply a single source of privileged information (PI) uniformly across all steps (top): on error steps such as $t_3$ and $t_5$, environment feedback (e.g., ``Nothing happens'') only signals failure and provides no corrective direction. \method (bottom) uses environment feedback as base PI on all steps, and adaptively provides an LLM-generated corrective hint on error steps, telling the agent what it should have done.}
  \label{fig:teaser}
\end{figure}

We observe that multi-turn environments naturally produce a step-specific signal that current methods overlook: environment feedback.
After each action, the environment returns an observation that is unavailable to the student when generating its response, making it a natural source of PI\@.
For routine steps, this feedback is sufficient: signals such as ``You arrived at shelf~2'' confirm the action's outcome, and the teacher's assessment shifts only mildly.
For error steps, environment feedback is necessary but insufficient: signals such as ``Nothing happens'' indicate failure but reveal neither the cause nor the corrective action.
LLM-generated corrective hints (e.g., ``Go to diningtable~1 to find the target object'') supply the missing piece: what the agent should have done at this specific step.
Together, the two sources produce naturally adaptive supervision: weak confirmatory signals on routine steps and strong corrective signals on error steps, without any explicit gating or per-step coefficient.

Based on this observation, we propose \method (\methodexp) for multi-turn agentic RL\@.
For each failed trajectory, \method injects environment feedback into the teacher context at every step, and additionally provides LLM-generated corrective hints at error steps identified by an LLM analyzer.
Successful trajectories, where the GRPO advantage already provides the correct gradient direction, bypass the PI pipeline and retain the vanilla advantage.
We use the token-level distillation signal to reweight the GRPO advantage following~\citet{rlsd}, which requires minimal changes to the standard algorithm.
At inference time, no PI, LLM calls, or environment feedback injection are needed.

We validate \method across the Qwen2.5~\citep{yang2024qwen} and Qwen3~\citep{yang2025qwen} model families on three benchmarks for LLM-based agents: ALFWorld~\citep{shridhar2020alfworld}, WebShop~\citep{yao2022webshop}, and Search-based QA~\citep{jin2025searchr1}.
\method achieves substantial improvements over GRPO ($+13.3$ points on ALFWorld and $+11.0$ on WebShop-Succ at 7B) and consistently outperforms self-distillation baselines such as SDAR~\citep{sdar}, Skill-SD~\citep{wang2026skillsd}, and RLSD~\citep{rlsd}. 

Our contributions are:
\begin{itemize}
\item We identify that routine and error steps in multi-turn trajectories require fundamentally different supervision, and that environment feedback combined with LLM corrective hints naturally provides on-policy, step-aware PI\@.
\item We propose \method, which constructs step-appropriate PI and selectively applies it to failed trajectories, integrating into GRPO with minimal changes.
\item We validate \method on three agentic benchmarks across three model scales, showing consistent improvements over GRPO and self-distillation baselines.
\end{itemize}

\section{Method}

\begin{figure}[t]
  \centering
  \includegraphics[width=\textwidth]{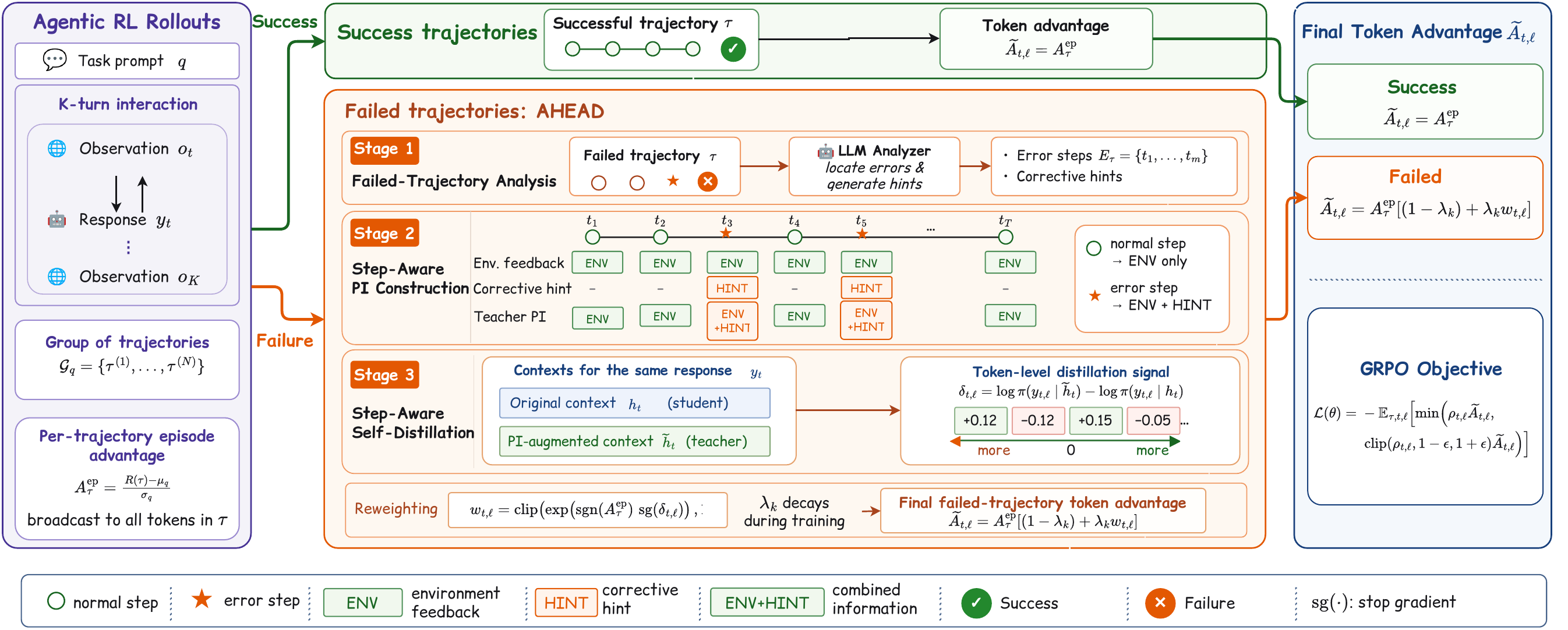}
  \caption{Overview of \method\@. \textbf{Stage~1} identifies error steps in failed trajectories via an LLM analyzer. \textbf{Stage~2} constructs step-aware privileged information: environment feedback for routine steps, environment feedback combined with an LLM corrective hint for error steps. \textbf{Stage~3} computes a token-level self-distillation signal $\delta_{t,\ell}$ by comparing the policy's log-probabilities under original and PI-augmented contexts, and converts it into a bounded reweight of the GRPO advantage. Successful trajectories bypass the PI pipeline entirely and retain the vanilla GRPO advantage.}
  \label{fig:main}
  \vspace{-10pt}
\end{figure}

We present \method (\textbf{A}daptive \textbf{H}indsight with \textbf{E}nvironment-\textbf{A}ugmented \textbf{D}istillation), a framework that constructs step-aware privileged information for multi-turn agent self-distillation and integrates the resulting token-level signal into the GRPO objective.
Figure~\ref{fig:main} illustrates the overall pipeline.

\subsection{Preliminaries}

\paragraph{Problem Setting.}
We consider a multi-turn setting, in which an agent interacts with an environment over a finite horizon.
At step $t$, the agent receives an observation $o_t$ and maintains an interaction history $h_t = (o_0, y_0, o_1, y_1, \ldots, o_t)$, where $y_i$ denotes the response generated at step~$i$.
The policy $\pi_\theta$ generates the next response as $y_t \sim \pi_\theta(\cdot \mid h_t)$.
A completed trajectory is $\tau = \{(o_t, y_t)\}_{t=0}^{T-1}$, with terminal outcome reward $R(\tau)$.

\paragraph{GRPO.}
For each task prompt $q$, GRPO samples $N$ trajectories $\mathcal{G}_q = \{\tau^{(1)}, \ldots, \tau^{(N)}\}$ and computes a group-relative advantage:
\begin{equation}
A^{\mathrm{ep}} = \frac{R(\tau) - \mu_q}{\sigma_q},
\end{equation}
where $\mu_q$ and $\sigma_q$ are the group mean and standard deviation.
This scalar is broadcast to every token in the trajectory.
The policy is optimized with the clipped surrogate:

\begin{equation}
\mathcal{L}_{\mathrm{GRPO}}(\theta) = -\mathbb{E}_{\tau,t,\ell}\!\left[\min\!\left(\rho_{t,\ell}\, A^{\mathrm{ep}},\; \hat{\rho}_{t,\ell}\, A^{\mathrm{ep}}\right)\right],
\end{equation}

where $\rho_{t,\ell} = \pi_\theta(y_{t,\ell} \mid h_t, y_{t,<\ell}) \,/\, \pi_{\theta_{\mathrm{old}}}(y_{t,\ell} \mid h_t, y_{t,<\ell})$ is the token-level importance ratio and $\hat{\rho}_{t,\ell} = \mathrm{clip}(\rho_{t,\ell},\, 1{-}\epsilon,\, 1{+}\epsilon)$ is its clipped counterpart.
Since $A^{\mathrm{ep}}$ is identical for every token, GRPO provides no mechanism to distinguish which steps or tokens were responsible for the outcome.

\paragraph{On-Policy Self-Distillation.}
On-policy self-distillation~\citep{rlsd} uses the same model as both teacher and student: the student scores each token under the original context $h_t$, while the teacher receives privileged information (PI) unavailable at decision time. The log-probability gap

\begin{equation}
\delta_{t,\ell} = \log \pi_{\theta_{\mathrm{old}}}(y_{t,\ell} \mid \tilde{h}_t, y_{t,<\ell}) - \log \pi_{\theta_{\mathrm{old}}}(y_{t,\ell} \mid h_t, y_{t,<\ell})
\label{eq:delta}
\end{equation}

measures how the PI revises the model's assessment of each sampled token.
If $\delta_{t,\ell} > 0$, the PI-augmented teacher assigns higher probability to the token than the student, endorsing the student's choice; if $\delta_{t,\ell} < 0$, the teacher disfavors it, suggesting the token should be suppressed.
The informativeness of $\delta$ depends entirely on what PI is injected into $\tilde{h}_t$, which is the focus of our method.

\subsection{Step-Aware Privileged Information}
\label{sec:pi}

\method constructs different PI for routine and error steps.
We first describe how error steps are identified, then how the two PI sources are combined into a step-aware teacher context.

\paragraph{Error Step Identification.}
After a trajectory completes with a failure outcome, the full trajectory record (observations, actions, environment feedback, and terminal outcome) is passed to an LLM-based analyzer.
The analyzer identifies steps whose actions were critical errors (e.g., picking up the wrong object, navigating to an irrelevant location).
We denote the set of identified error steps as $\mathcal{E}_\tau$.

\paragraph{Environment Feedback as PI.}
After the agent generate an action at step $t$, the environment returns an observation $o_{t+1}$.
This feedback is not available to the student when generating $y_t$, making it a natural source of PI\@.
For routine steps, it confirms that the action was appropriate (e.g., ``You pick up the mug from shelf~2'').
For error steps, it signals that something went wrong (e.g., ``Nothing happens'').
Environment feedback is grounded and local, reflecting the actual consequence of the agent's specific action. However, it only describes \emph{what happened}, not \emph{what should have happened}.

\paragraph{LLM Corrective Hints as PI.}
For each error step $t \in \mathcal{E}_\tau$, the LLM analyzer generates a corrective hint describing what the agent should have done instead.
For example, if the agent attempted to pick up an object from a wrong location, the hint might state: ``You should first go to diningtable~1 to find the target object.''
While environment feedback can only signal that an action failed, the corrective hint provides the missing direction: \emph{why} the action was wrong and \emph{what} the alternative should be.

\paragraph{PI Construction.}We construct the PI-augmented context $\tilde{h}_t$ by injecting the appropriate PI:
\begin{equation}
\tilde{h}_t = \begin{cases}
H(h_t,\, \Phi^{\mathrm{env}}_t,\, \Phi^{\mathrm{llm}}_t) & t \in \mathcal{E}_\tau \\[2pt]
H(h_t,\, \Phi^{\mathrm{env}}_t) & t \notin \mathcal{E}_\tau
\end{cases}
\label{eq:pi_injection}
\end{equation}
where $H(\cdot)$ appends PI to the history, $\Phi^{\mathrm{env}}_t$ is the environment feedback, and $\Phi^{\mathrm{llm}}_t$ is the corrective hint.
Error steps receive both sources simultaneously: the feedback identifies the failure, and the hint supplies the correction.
Because the teacher sees richer PI on error steps, the resulting $|\delta_{t,\ell}|$ is naturally larger than on routine steps, producing stronger token-level signals on error steps without any explicit gating or per-step coefficient.
Note that the LLM analyzer is invoked only on failed trajectories, while environment feedback requires no additional computation and covers all steps.

\subsection{Selective Trajectory Filtering}
\label{sec:filter}

Not all trajectories benefit equally from PI-based supervision.
Failed trajectories carry a uniform negative advantage but lack information about which steps were responsible. This is precisely the gap that the self-distillation signal $\delta$ can fill.
Successful trajectories already carry positive advantages that provide the correct gradient direction; injecting PI risks introducing noise when the feedback is not aligned with the tokens that led to success.

\method therefore applies PI-based reweighting only to failed trajectories.
Let $\mathcal{M}_\tau$ denote the set of steps that participate in reweighting:
\begin{equation}
\mathcal{M}_\tau =
\begin{cases}
\{\, t : \Phi_t \neq \emptyset \,\} & \tau \text{ failed} \\[2pt]
\emptyset & \tau \text{ succeeded}
\end{cases}
\label{eq:filter}
\end{equation}
where $\Phi_t$ denotes the PI available at step~$t$.
For steps outside $\mathcal{M}_\tau$, the reweighted advantage in Sec.~\ref{sec:reweight} reduces to $A^{\mathrm{ep}}$.

\subsection{Advantage Reweighting}
\label{sec:reweight}

Following RLSD~\citep{rlsd}, we convert the token-level self-distillation signal into a multiplicative weight on the GRPO advantage, rather than using it as a separate distillation loss.
This ensures that the environment reward determines whether the policy is reinforced or penalized, while the distillation signal only adjusts how strongly each token is updated.

Specifically, we exponentiate the gap, gated by the sign of the episode advantage:
\begin{equation}
w_{t,\ell} = \mathrm{clip}\big( \exp\!\big(\mathrm{sgn}(A^{\mathrm{ep}})\, \mathrm{sg}(\delta_{t,\ell})\big),\; 1{-}\varepsilon,\; 1{+}\varepsilon \big)
\label{eq:weight}
\end{equation}
\begin{equation}
\tilde{A}_{t,\ell} = A^{\mathrm{ep}} \cdot \big[(1-\lambda_k) + \lambda_k \cdot w_{t,\ell}\big]
\label{eq:reweight}
\end{equation}
where $\mathrm{sg}$ denotes stop-gradient and $\varepsilon = 0.2$.
Since \method applies reweighting only to failed trajectories (Sec.~\ref{sec:filter}), where $A^{\mathrm{ep}} < 0$, the effect is straightforward: tokens disfavored by the teacher ($\delta < 0$) receive larger penalties, while tokens endorsed by the teacher ($\delta > 0$) receive smaller penalties.
Because $\exp(\cdot) > 0$ and clipping preserves positivity, the reweighted advantage always preserves $\mathrm{sign}(\tilde{A}) = \mathrm{sign}(A^{\mathrm{ep}})$, so the environment reward always controls the update direction.

The mixing coefficient $\lambda_k$ decays linearly from $\lambda_0 = 0.5$ to $0$ over $D$ training steps.
Early training benefits from dense PI-based credit assignment, while later training returns to vanilla GRPO to avoid over-reliance on privileged information.

\paragraph{Objective.}
The final objective replaces $A^{\mathrm{ep}}$ with $\tilde{A}_{t,\ell}$ in the GRPO clipped surrogate:
\begin{equation}
\mathcal{L}(\theta) = -\mathbb{E}_{\tau,t,\ell}\!\left[\min\!\left(\rho_{t,\ell}\, \tilde{A}_{t,\ell},\; \hat{\rho}_{t,\ell}\, \tilde{A}_{t,\ell}\right)\right].
\end{equation}

\paragraph{Training--Inference Boundary.}
The LLM analyzer, corrective hints, and PI-augmented scoring are used only during training to construct the advantage.
At inference time, the policy acts from $h_t$ alone, without any PI, LLM calls, or environment feedback injection.

\section{Experiments}

\subsection{Experimental Setting}

\paragraph{Benchmarks.}
We conduct experiments on three agentic benchmarks that cover embodied reasoning, web navigation, and search-augmented question answering.

ALFWorld \citep{shridhar2020alfworld} is a text-based household environment where an agent must complete language-specified goals via sequential textual actions. It includes six task categories: Pick, Look, Clean, Heat, Cool, and Pick2. We use the training split from GiGPO \citep{feng2025gigpo}.

WebShop \citep{yao2022webshop} simulates an e-commerce website where an agent searches for and purchases products matching natural-language specifications. We evaluate on 128 fixed tasks following prior work \citep{feng2025gigpo}, reporting both task-completion score and binary success rate.

Search-based QA \citep{jin2025searchr1} requires an agent to answer questions by issuing search queries and reading returned documents. The benchmark spans single-hop (NQ~\citep{kwiatkowski2019nq}, TriviaQA~\citep{joshi2017triviaqa}, PopQA~\citep{mallen2023popqa}) and multi-hop (HotpotQA~\citep{yang2018hotpotqa}, 2WikiMultiHopQA~\citep{ho2020twowiki}, MuSiQue~\citep{trivedi2022musique}, Bamboogle~\citep{press2023bamboogle}) datasets. Following Search-R1~\citep{jin2025searchr1}, we train on NQ and HotpotQA; the remaining datasets serve as out-of-domain evaluation. Full dataset and metric details are given in Appendix~\ref{app:datasets}.

\paragraph{Baselines.}
We compare against three categories of methods.
(1) Training-free: Vanilla uses the base instruction-tuned model without post-training. Skill-Prompt* retrieves a task-relevant skill and prepends it to the prompt at inference time.
(2) RL methods: GRPO \citep{grpo} optimizes the policy with group-relative trajectory-level advantages. Skill-GRPO augments GRPO by injecting retrieved skills into training prompts; Skill-GRPO* additionally retains skills at validation time.
(3) Hybrid methods combine RL with self-distillation or skill-conditioned supervision. OPSD \citep{opsd} distills token-level knowledge from a frozen reference policy. GRPO+OPSD adds OPSD as an auxiliary loss on top of GRPO. Skill-SD \citep{wang2026skillsd} uses importance-weighted distillation with retrieved skills as privileged context. RLSD \citep{rlsd} re-weights GRPO advantages using the teacher-student log-probability gap. SDAR \citep{sdar} applies a sigmoid-gated auxiliary distillation loss that selectively distills teacher-endorsed tokens.

\paragraph{Implementation Details.}
We use Qwen2.5-3B/7B-Instruct \citep{yang2024qwen} and Qwen3-1.7B-Instruct \citep{yang2025qwen} as backbone models. All models are trained for 150 steps on 8 H100 GPUs. For ALFWorld and WebShop, each batch samples 16 tasks with 8 rollouts per prompt. For Search-based QA, the batch size is 128 tasks. The maximum prompt length is 2,048 for ALFWorld and 4,096 for WebShop and Search-based QA. We use a GRPO clipping range of $\epsilon = 0.2$ and a reweight bound of $\varepsilon = 0.2$; the distillation coefficient decays linearly from $\lambda_0 = 0.5$ to $0$ over $D = 50$ training steps (Sec.~\ref{sec:reweight}). Error detection and corrective hint generation use Claude Opus~4.7 as the LLM analyzer. Full hyperparameters are provided in Appendix~\ref{app:hyper}.

\paragraph{Evaluation Protocol.}
We report ALFWorld results on \textbf{Val-128}: a fixed 128-task set sampled from the seen split, following \citet{sdar}. We evaluate generalization to unseen room layouts on the \textbf{Unseen-134} split. Results on the full \textbf{Seen-140} and \textbf{Unseen-134} splits are provided in Appendix~\ref{app:results}.

\begin{table}[t!]
    \centering
    \resizebox{1\textwidth}{!}{%
    \begin{tabular}{l ccccccc cccccccc cc}
    \toprule
    & \multicolumn{7}{c}{\textbf{ALFWorld}} & \multicolumn{8}{c}{\textbf{Search-based QA}} & \multicolumn{2}{c}{\textbf{WebShop}} \\
    \cmidrule(lr){2-8} \cmidrule(lr){9-16} \cmidrule(lr){17-18}
    \textbf{Method}
    & \textbf{Pick} & \textbf{Look} & \textbf{Clean} & \textbf{Heat} & \textbf{Cool} & \textbf{Pick2} & \textbf{Avg}
    & \textbf{NQ} & \textbf{Triv} & \textbf{Pop} & \textbf{Hotp} & \textbf{2Wk} & \textbf{MuS} & \textbf{Bam} & \textbf{Avg}
    & \textbf{Score} & \textbf{Succ.} \\
    \midrule
    \rowcolor{gray!10} \multicolumn{18}{l}{\textit{Qwen2.5-3B-Instruct}} \\

    Vanilla
        & 44.4 & 11.1 & 6.2 & 15.4 & 28.6 & 12.5 & 21.9
        & 24.6 & 48.1 & 31.0 & 26.3 & 25.3 & 7.2 & 59.7 & 31.7
        & 6.7 & 0.8
        \\

    Skill-Prompt*
        & 51.7 & 66.7 & 48.4 & 0.0 & 4.3 & 10.0 & 28.9
        & 23.7 & 46.2 & 30.6 & 24.4 & 22.1 & 7.5 & 12.5 & 23.9
        & 0.2 & 0.8
        \\

    OPSD
        & 48.8 & 41.7 & 16.7 & 0.0 & 15.8 & 16.7 & 28.1
        & 0.1 & 0.1 & 0.1 & 0.0 & 0.0 & 0.0 & 0.0 & 0.0
        & 11.3 & 3.1
        \\

    GRPO
        & 91.2 & 62.5 & 96.2 & 61.9 & 65.0 & 47.4 & 75.0
        & 39.3 & \underline{60.6} & 41.1 & 37.4 & 34.6 & 15.4 & 26.4 & 36.4
        & 79.8 & 63.3
        \\

    Skill-GRPO
        & 88.9 & 71.4 & 58.8 & 70.6 & 40.7 & 29.2 & 60.2
        & 43.5 & 58.8 & 43.0 & 36.8 & 32.2 & 11.7 & 12.5 & 34.1
        & 77.3 & 60.9
        \\

    Skill-GRPO*
        & 94.3 & 57.1 & \textbf{100.0} & 66.7 & \underline{73.1} & 57.1 & 80.5
        & 44.3 & 59.6 & 44.3 & 39.0 & 36.1 & 14.5 & 14.9 & 36.1
        & 76.3 & 66.4
        \\

    GRPO+OPSD
        & \textbf{100.0} & \textbf{82.4} & 85.7 & \underline{75.0} & 70.0 & 60.0 & 81.2
        & \underline{44.9} & \textbf{61.2} & \underline{45.2} & \textbf{40.4} & 38.5 & \underline{16.0} & \underline{66.1} & \textbf{44.6}
        & 77.8 & 66.4
        \\

    Skill-SD
        & 88.2 & 50.0 & 96.2 & 52.4 & 65.0 & 57.9 & 73.4
        & 44.4 & 60.4 & 44.0 & 39.5 & \underline{40.4} & 15.4 & 64.9 & 44.1
        & 75.9 & 64.0
        \\

    RLSD
        & 87.9 & \underline{75.0} & 90.9 & \underline{75.0} & \underline{73.1} & 68.4 & 79.7
        & 41.5 & 58.6 & 42.3 & \textbf{40.4} & 40.2 & \textbf{16.8} & \textbf{66.9} & 43.8
        & 84.4 & 66.4
        \\

    SDAR
        & \underline{97.1} & 62.5 & \textbf{100.0} & 61.9 & \textbf{75.0} & \underline{84.2} & \underline{84.4}
        & 44.8 & 58.1 & 44.3 & 38.6 & 36.2 & 15.7 & \underline{66.1} & 43.4
        & \underline{85.0} & \underline{68.0}
        \\

    \rowcolor{oursrow} \textbf{\method}
        & \underline{97.1} & 54.5 & \underline{96.7} & \textbf{90.9} & 72.7 & \textbf{89.5} & \textbf{87.5}
        & \textbf{46.1} & \textbf{61.2} & \textbf{49.2} & \underline{40.3} & \textbf{41.7} & 12.7 & 59.2 & \underline{44.3}
        & \textbf{88.5} & \textbf{73.4}
        \\

    \midrule

    \rowcolor{gray!10} \multicolumn{18}{l}{\textit{Qwen2.5-7B-Instruct}} \\

    Vanilla
        & 36.1 & 22.2 & 3.1 & 0.0 & 0.0 & 0.0 & 12.5
        & 25.2 & 50.8 & 29.5 & 29.0 & 29.0 & 10.4 & 63.7 & 33.9
        & 5.9 & 1.6
        \\

    Skill-Prompt*
        & 51.7 & 50.0 & 32.3 & 5.3 & 4.3 & 0.0 & 23.4
        & 30.9 & 52.1 & 32.7 & 32.7 & 27.9 & 12.7 & 66.1 & 36.4
        & 1.7 & 0.8
        \\

    OPSD
        & 50.0 & 60.0 & 22.7 & 21.4 & 17.6 & 9.5 & 32.8
        & 8.8 & 8.6 & 17.5 & 2.5 & 4.2 & 0.5 & 1.2 & 6.2
        & 4.5 & 2.3
        \\

    GRPO
        & 91.2 & \underline{87.5} & 96.2 & 81.0 & 65.0 & 57.9 & 81.2
        & 45.1 & 63.7 & 44.0 & 43.6 & 43.2 & 16.8 & 37.6 & 42.0
        & 80.9 & 72.6
        \\

    Skill-GRPO
        & 88.5 & 66.7 & 65.2 & 61.1 & 57.7 & 73.1 & 69.5
        & 45.2 & 63.7 & 45.7 & 43.1 & 43.3 & 19.6 & 21.4 & 40.3
        & 80.4 & 71.9
        \\

    Skill-GRPO*
        & \textbf{100.0} & 83.3 & \underline{96.4} & 83.3 & 75.0 & \underline{78.9} & \underline{88.3}
        & 44.8 & 63.0 & 45.1 & 43.7 & 43.7 & \underline{20.5} & \underline{71.4} & 47.5
        & 87.0 & 81.2
        \\

    GRPO+OPSD
        & 91.4 & 61.5 & \textbf{100.0} & \underline{87.5} & 76.5 & 52.2 & 80.4
        & \textbf{47.3} & \underline{64.5} & 46.9 & 43.8 & 39.3 & 18.0 & 69.4 & 47.0
        & 86.8 & 76.5
        \\

    Skill-SD
        & 93.9 & \textbf{93.8} & 90.9 & \textbf{100.0} & 69.2 & 68.4 & 85.1
        & \underline{47.1} & \underline{64.5} & 47.8 & \underline{44.2} & 42.1 & 20.2 & 69.0 & 47.8
        & 86.1 & 76.5
        \\

    RLSD
        & \textbf{100.0} & \underline{87.5} & 92.3 & 58.8 & \underline{80.0} & 65.2 & 82.0
        & 46.8 & 63.0 & 44.4 & \textbf{45.5} & \textbf{48.9} & \textbf{21.5} & \textbf{73.0} & \textbf{49.0}
        & 87.4 & 77.3
        \\

    SDAR
        & 94.7 & 75.0 & \textbf{100.0} & 86.7 & 68.2 & \underline{78.9} & 85.9
        & 46.3 & 63.5 & \underline{48.2} & 43.8 & \underline{48.4} & 19.6 & \textbf{73.0} & \textbf{49.0}
        & \underline{89.4} & \underline{82.8}
        \\

    \rowcolor{oursrow} \textbf{\method}
        & \underline{96.7} & 81.8 & \textbf{100.0} & \textbf{100.0} & \textbf{89.3} & \textbf{95.0} & \textbf{94.5}
        & \underline{47.1} & \textbf{64.7} & \textbf{48.3} & \textbf{45.5} & 44.7 & 19.3 & 69.8 & \underline{48.5}
        & \textbf{89.9} & \textbf{83.6}
        \\

    \midrule

    \rowcolor{gray!10} \multicolumn{18}{l}{\textit{Qwen3-1.7B-Instruct}} \\

    Vanilla
        & 25.0 & 22.2 & 3.1 & 0.0 & 21.4 & 4.2 & 12.5
        & 29.4 & 46.9 & 37.0 & 23.5 & 19.6 & 6.4 & 10.5 & 24.8
        & 46.5 & 4.7
        \\

    Skill-Prompt*
        & 10.3 & 50.0 & 16.1 & 0.0 & 0.0 & 5.0 & 9.4
        & 29.4 & 46.5 & 36.2 & 22.9 & 20.8 & 4.3 & 10.1 & 24.3
        & 23.0 & 2.3
        \\

    OPSD
        & 26.3 & 33.3 & 9.1 & 0.0 & 4.5 & 5.3 & 14.1
        & 4.2 & 8.3 & 4.6 & 6.6 & 15.3 & 0.7 & 1.2 & 5.8
        & 47.4 & 9.3
        \\

    GRPO
        & 71.1 & 41.7 & 36.4 & 40.0 & 31.8 & 31.6 & 46.1
        & 40.0 & \underline{58.9} & 43.5 & 35.4 & 30.3 & 12.0 & 65.7 & 40.8
        & 67.3 & 38.3
        \\

    Skill-GRPO
        & 27.6 & \underline{54.5} & 22.7 & 27.3 & 0.0 & 19.2 & 21.1
        & 39.2 & 58.6 & 43.9 & 35.2 & 28.2 & 11.5 & \underline{66.1} & 40.4
        & 73.4 & 46.1
        \\

    Skill-GRPO*
        & 31.4 & 42.9 & 51.9 & 8.3 & 11.5 & 7.1 & 28.1
        & 38.0 & 58.4 & 43.9 & 36.3 & 29.0 & 12.5 & \textbf{66.9} & 40.7
        & \underline{80.4} & 50.0
        \\

    GRPO+OPSD
        & 38.2 & 50.0 & 30.8 & 28.6 & 30.0 & 21.1 & 32.0
        & \underline{40.7} & \underline{58.9} & 45.0 & \underline{37.0} & 34.6 & \textbf{13.3} & 65.7 & \textbf{42.2}
        & 70.7 & 38.3
        \\

    Skill-SD
        & 52.9 & 37.5 & \underline{69.2} & \underline{42.9} & \underline{60.0} & \underline{36.8} & 52.3
        & 39.1 & 57.5 & \underline{45.4} & 34.8 & 34.1 & 10.7 & 64.1 & 40.8
        & \textbf{81.8} & 53.9
        \\

    RLSD
        & 50.0 & 37.5 & 61.5 & 19.0 & 50.0 & 21.1 & 42.2
        & 38.6 & 57.3 & 43.0 & 34.5 & 34.1 & 11.5 & 65.3 & 40.6
        & 74.0 & 50.8
        \\

    SDAR
        & \textbf{73.5} & 25.0 & \textbf{76.9} & 33.3 & 40.0 & \underline{36.8} & \underline{53.9}
        & 39.7 & \underline{58.9} & 45.3 & 35.9 & \underline{35.5} & \underline{12.6} & 65.3 & \underline{41.9}
        & 76.8 & \underline{58.6}
        \\

    \rowcolor{oursrow} \textbf{\method}
        & \underline{73.3} & \textbf{63.6} & 63.6 & \textbf{52.9} & \textbf{71.4} & \textbf{70.0} & \textbf{67.2}
        & \textbf{43.3} & \textbf{59.0} & \textbf{48.2} & \textbf{39.0} & \textbf{36.8} & 10.9 & 56.0 & \underline{41.9}
        & 79.3 & \textbf{64.8}
        \\

    \bottomrule
    \end{tabular}
    }
    \vspace{8pt}
    \caption{
       \textbf{Performance Comparison on long-horizon benchmarks.} We report the success rate (\%) on ALFWorld, accuracy on search-based QA, and task-completion score/success rate on WebShop. An asterisk (*) denotes validation with skills. Within each backbone block, the \textbf{best} result in a column is in bold and the \underline{second-best} is underlined. Baseline numbers are taken from \citet{sdar}.
    }
    \label{tab:main_results}
\end{table}

\subsection{Main Results}

\paragraph{Overall Performance.}
\method achieves the best or second-best result in most columns in Table~\ref{tab:main_results}. Compared to GRPO, it delivers consistent gains on ALFWorld ($+12.5$ points on 3B, $+13.3$ on 7B, $+21.1$ on 1.7B) and WebShop-Succ ($+10.1$, $+11.0$ and $+26.5$ points respectively), and raises Search-Avg at every scale ($+7.9$, $+6.5$ and $+1.1$). On the two interactive benchmarks the improvements are most pronounced on the smaller Qwen3-1.7B. This is expected: smaller models make more errors per trajectory, so there are more error steps where corrective hints can provide useful supervision.

\paragraph{Comparison with Self-Distillation Baselines.}
On ALFWorld, \method matches or outperforms all self-distillation baselines across every model scale. At 7B, it reaches 94.5\%, surpassing SDAR (85.9\%), Skill-SD (85.1\%), and RLSD (82.0\%) by large margins. At 1.7B, where smaller models struggle to utilize retrieved skills effectively, \method achieves 67.2\% while SDAR reaches 53.9\% and RLSD only 42.2\%. On WebShop, \method leads across all three model scales, reaching 83.6\% Succ at 7B against SDAR's 82.8\%.

\paragraph{Stability.}
Standalone OPSD collapses catastrophically (near-zero on Search-QA), and the naive GRPO+OPSD combination degrades severely on Qwen3-1.7B (32.0\% vs.\ 46.1\% for GRPO on ALFWorld) due to unbounded distillation gradients overwhelming the RL signal. \method avoids these instabilities entirely, because the reweighting (Sec.~\ref{sec:reweight}) can only adjust how strongly each token's advantage is applied, never its sign.

\subsection{Analysis}
\label{sec:analysis}

\begin{figure}[t]
  \centering
  \begin{minipage}[t]{0.49\textwidth}
    \centering
    \includegraphics[width=\linewidth]{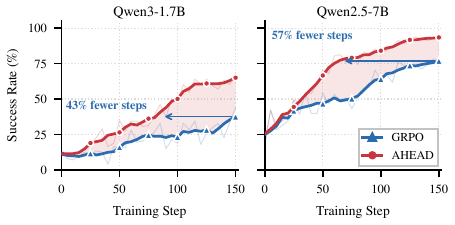}
    \caption{Sample efficiency on ALFWorld, measured on the seen split during training, for Qwen3-1.7B (left) and Qwen2.5-7B (right). \method converges faster and reaches a higher final success rate than GRPO at both scales; arrows mark the training-step saving to reach GRPO's final score.}
    \label{fig:sample_eff}
  \end{minipage}\hfill
  \begin{minipage}[t]{0.49\textwidth}
    \centering
    \includegraphics[width=\linewidth]{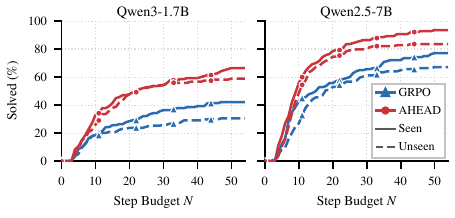}
    \caption{
    Fraction of ALFWorld tasks solved within N steps for Qwen3-1.7B (left) and Qwen2.5-7B (right). Solid and dashed lines denote seen and unseen splits, respectively. Beyond the first few steps ($N \geq 5$), \method consistently outperforms GRPO across both splits and scales.}
    \label{fig:traj_cdf}
  \end{minipage}
  \vspace{-10pt}
\end{figure}

\paragraph{Sample Efficiency.}
As shown in Figure~\ref{fig:sample_eff}, \method converges substantially faster than GRPO. It reaches GRPO's final performance level early in training and continues to improve to $94.5$ against GRPO's $81.2$. The same pattern holds at 1.7B, where \method ends $21.1$ points above GRPO ($67.2$ vs.\ $46.1$). Step-aware PI therefore does not merely raise the final score; it reaches any given score sooner, because the corrective hints tell the policy what it should have done at each error step rather than leaving it to infer this from trajectory-level rewards alone.

\paragraph{Step Efficiency at Test Time.}
The efficiency gains transfer to test time (Figure~\ref{fig:traj_cdf}). Beyond the first few steps ($N \geq 5$), \method's curve lies above GRPO's on both the seen and unseen splits of ALFWorld, so the improvement is not an artifact of a generous interaction limit. The gap is largest at practical step budgets: on the unseen split, \method solves 74.6\% of tasks within 20 steps where GRPO solves 53.0\%, and \method needs only 15 steps to match the success rate GRPO attains with its full 54-step budget. The margin is wider at 1.7B, where \method solves 47.8\% of tasks within 20 steps against GRPO's 23.9\% and needs only 12 steps to reach GRPO's full-budget coverage. The gains therefore come from eliminating unnecessary steps, rather than needing more room to explore. In deployment, this usually means lower latency and fewer API calls, not merely a higher score.

\paragraph{Ablation Study.}
%
\newcommand{\yes}{\checkmark}
\newcommand{\no}{\textcolor{gray!70}{$\times$}}
\newcommand{\na}{\textcolor{gray!50}{--}}

\begin{table}[t]
    \centering
    \small
    \setlength{\tabcolsep}{4.5pt}
    \resizebox{\textwidth}{!}{%
    \begin{tabular}{l ccccc ccc ccc}
    \toprule
    & \multicolumn{5}{c}{\textbf{Components}}
    & \multicolumn{3}{c}{\textbf{ALFWorld}} & \multicolumn{3}{c}{\textbf{WebShop}} \\
    \cmidrule(lr){2-6} \cmidrule(lr){7-9} \cmidrule(lr){10-12}
    \textbf{Configuration}
    & \textbf{Env.\ fb.} & \textbf{Hints} & \textbf{Multi} & \textbf{Fail-only} & \textbf{Decay}
    & \textbf{1.7B} & \textbf{3B} & \textbf{7B}
    & \textbf{1.7B} & \textbf{3B} & \textbf{7B} \\
    \midrule
    \rowcolor{oursrow} \textbf{\method}
        & \yes & \yes & \yes & \yes & \yes
        & \textbf{67.2} & \textbf{87.5} & \textbf{94.5}
        & 64.8 & \textbf{73.4} & \textbf{83.6} \\
    \midrule
    w/o decay
        & \yes & \yes & \yes & \yes & \no
        & 60.2 & 71.1 & 85.2 & 54.7 & 66.4 & \textbf{83.6} \\
    w/o failure-only
        & \yes & \yes & \yes & \no & \yes
        & 48.4 & 82.0 & 93.0 & 64.1 & 72.7 & 75.8 \\
    w/o multi-step
        & \yes & \yes & \no & \yes & \yes
        & 61.7 & 81.2 & 86.7 & 57.0 & 47.7 & 75.0 \\
    w/o hints
        & \yes & \no & \na & \yes & \yes
        & 53.1 & 73.4 & 87.5 & 63.3 & 68.0 & 74.2 \\
    w/o env.\ feedback
        & \no & \yes & \yes & \yes & \yes
        & 59.4 & 81.2 & 89.8 & \textbf{67.2} & 72.7 & 72.7 \\
    \midrule
    Env.\ feedback only
        & \yes & \no & \na & \no & \yes
        & 56.2 & 72.7 & 88.3
        & 60.2 & 71.9 & 78.9 \\
    \bottomrule
    \end{tabular}%
    }
    \vspace{8pt}
    \caption{
        \textbf{Component ablation.} Success rate (\%) on ALFWorld and WebShop. Ticks give the configuration of each run, so rows differing in a single tick are one-component comparisons. \emph{Env.\ fb.}: environment feedback as base PI\@. \emph{Hints}: the LLM-generated corrective hint $\Phi^{\mathrm{llm}}$ on error steps. \emph{Multi}: the analyzer may mark several error steps per trajectory rather than one. \emph{Fail-only}: PI is applied to failed trajectories only (Sec.~\ref{sec:filter}). \emph{Decay}: the linear schedule on $\lambda_k$. The last row removes both the hints and the failure-only filter, leaving environment feedback applied uniformly.
    }
    \label{tab:ablation}
\end{table}
Table~\ref{tab:ablation} removes one component at a time; the full method leads nearly every column.
Without the linear decay of $\lambda_k$, performance drops by $7.0$--$16.4$ points on ALFWorld and up to $10.1$ on WebShop, confirming that PI-based reweighting must fade as training progresses.
Applying PI to all trajectories instead of only failed ones hurts most at 1.7B ($-18.8$ on ALFWorld), where PI on a successful trajectory pulls against an already correct gradient.
Restricting the analyzer to a single error step per trajectory costs up to $25.7$ points (WebShop-3B), since failed episodes rarely contain just one mistake.
Both PI sources contribute: removing corrective hints costs $7.0$--$14.1$ points on ALFWorld, confirming that a failure signal alone is not enough without corrective direction; removing environment feedback incurs smaller but consistent losses on ALFWorld ($4.7$--$7.8$ points), though the relative importance of the two sources varies across benchmarks.

\paragraph{Choice of LLM Analyzer.}

\begin{wraptable}{r}{0.45\textwidth}
  \centering
  \small
  \vspace{-\intextsep}
  \begin{tabular}{lcc}
    \toprule
    \textbf{Analyzer} & \textbf{3B} & \textbf{1.7B} \\
    \midrule
    \rowcolor{oursrow} Opus 4.7 (default) & 87.5 & 67.2 \\
    Sonnet 5 & \textbf{89.1} & \textbf{68.0} \\
    Kimi & 83.6 & \textbf{68.0} \\
    GLM-5 & 81.2 & 67.2 \\
    \midrule
    GRPO (no analyzer) & 75.0 & 46.1 \\
    \bottomrule
  \end{tabular}
  \caption{ALFWorld success rate (\%) when the error-step analyzer is swapped, holding everything else fixed. 
  GRPO, which uses no analyzer, is repeated from Table~\ref{tab:main_results} as a floor.}
  \label{tab:analyzer}
  \vspace{-10pt}
\end{wraptable}

\method depends on an external LLM to select error steps and write corrective hints, so a natural question is how much of the gain is attributed to the analyzer. Table~\ref{tab:analyzer} swaps it for three alternatives while holding the rest of the pipeline fixed. Two observations follow. First, every analyzer beats GRPO by a wide margin (the weakest, GLM-5, still scores 81.2 at 3B against GRPO's 75.0, and 67.2 at 1.7B against 46.1), so the gain is not contingent on one particular model. Second, the spread across analyzers is scale-dependent: at 3B the four span 7.9 points (81.2--89.1), while at 1.7B they fall within 0.8 points of each other (67.2--68.0). At the smaller scale the policy appears unable to exploit the difference between a better and a worse hint, which puts a ceiling on what a stronger analyzer can buy: the bottleneck there is the policy's capacity to act on corrective information, not the quality of that information. A cheaper analyzer therefore suffices for smaller backbones.

\begin{figure}[t]
  \centering
  \includegraphics[width=\textwidth]{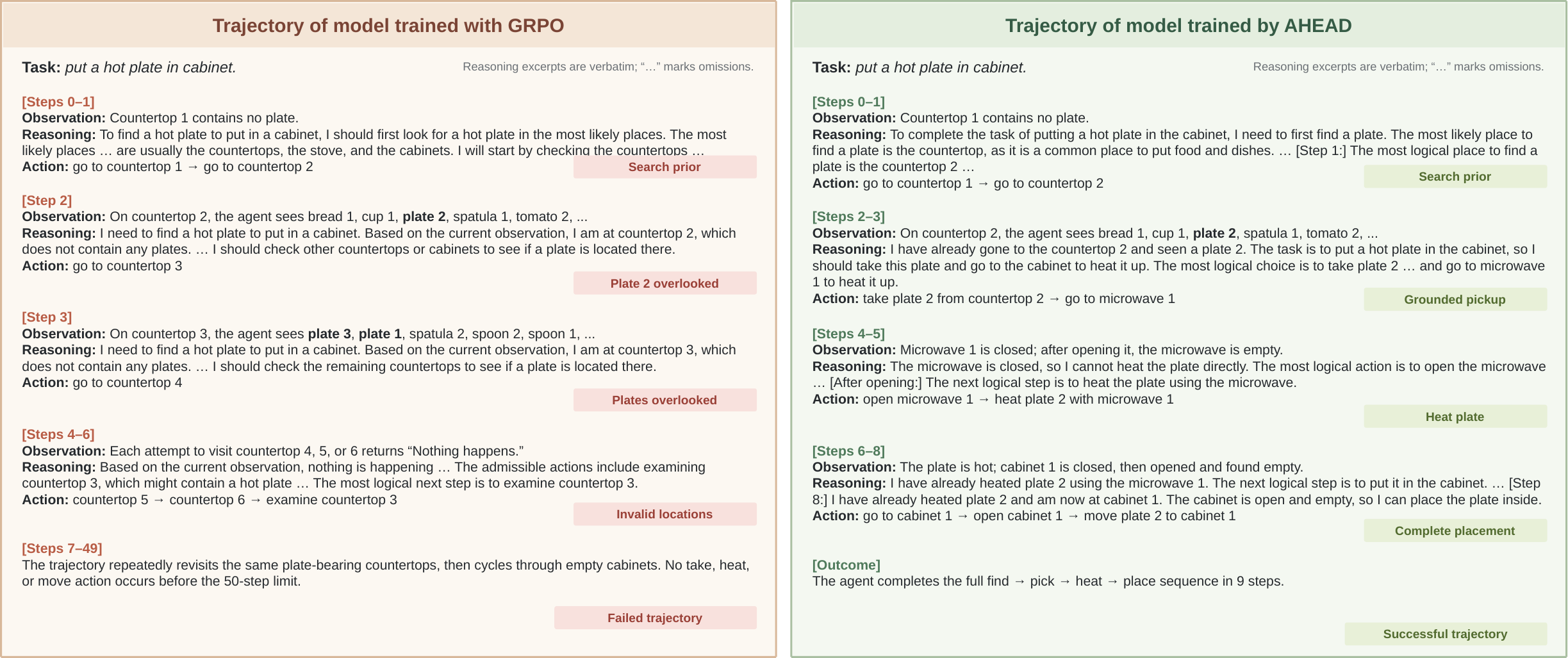}
  \caption{Trajectory comparison on an ALFWorld task \emph{put a hot plate in cabinet}. \textbf{Left:} a GRPO-trained policy overlooks the visible plates, revisits the same countertops and empty cabinets, and never takes, heats, or moves an object before the 50-step limit (failed trajectory). \textbf{Right:} a \method-trained policy grounds each action in the current observation and completes the find $\rightarrow$ pick $\rightarrow$ heat $\rightarrow$ place sequence in nine steps.
  Reasoning and observation excerpts are lightly condensed for space; ``$\ldots$'' marks omissions and retained text is verbatim.
  }
  \label{fig:trajectory}
  \vspace{-10pt}
\end{figure}

\paragraph{Qualitative Trajectory Comparison.}
Figure~\ref{fig:trajectory} contrasts a GRPO-trained policy with a \method-trained policy on the same ALFWorld task, \emph{put a hot plate in cabinet}. Both start from identical observations, but the two trajectories diverge immediately at the first plate-bearing countertop. The GRPO agent repeatedly overlooks the plates in front of it, cycles through the same countertops and empty cabinets, and never issues a single take, heat, or move action before hitting the 50-step limit, so the episode fails. The \method agent instead grounds each decision in the current observation: it picks up the plate it sees, heats it in the microwave, and places it in the cabinet, completing the full find $\rightarrow$ pick $\rightarrow$ heat $\rightarrow$ place sequence in nine steps. This is the failure mode that step-aware corrective hints target during training, an early perception error that GRPO compounds into an unrecoverable loop, and it is consistent with the shorter, more successful trajectories \method produces (Sec.~\ref{app:results}). The reasoning and observation text shown is lightly condensed for readability; ``$\ldots$'' marks omitted spans, and the retained excerpts are verbatim.

\section{Related Work}
\paragraph{Reinforcement Learning for LLM Agents.}
Reinforcement learning is now widely used for post-training language model agents in interactive environments \citep{grpo,dong2025arpo,feng2025gigpo}. Agents trained with RL interact with environments over many steps, making sequential decisions in settings such as embodied reasoning \citep{shridhar2020alfworld}, web navigation \citep{yao2022webshop}, and GUI automation \citep{lu2025uis1,lu2026uir1}. A central difficulty in these settings is credit assignment: outcome rewards indicate whether an episode succeeded but provide no information about which intermediate decisions were responsible \citep{deng2025energybasedtransferreinforcementlearning}.

\paragraph{On-Policy Self-Distillation for Agents.}
On-policy self-distillation provides an alternative source of dense supervision by letting the same model serve as both student and teacher under different contexts \citep{agarwal2024gkd,opsd,he2026sdzero}. Several recent methods combine this idea with RL for multi-turn agents. SDAR \citep{sdar} uses a sigmoid gate on the teacher-student log-probability gap to selectively distill teacher-endorsed tokens while attenuating noisy signals. RLSD \citep{rlsd} re-weights GRPO advantages using the same gap without a separate distillation loss. Skill-SD \citep{wang2026skillsd} augments the teacher with retrieved natural-language skills and applies importance-weighted distillation. These methods apply privileged information uniformly across steps or select among different granularities of the same PI source. Our work differs by using two heterogeneous PI sources and allocating them based on step type.

\paragraph{Privileged Information in Agent Training.}
Using PI during training while removing it at inference has roots in the learning-by-cheating paradigm \citep{chen2019cheating}. In LLM agent training, this idea appears as skill-conditioned learning, where natural-language skills are provided during training but removed at test time \citep{lu2026skill0,xia2026skillrl,shi2026skill1}. These approaches typically use a single type of privileged information applied uniformly across steps. Our work extends this line by combining multiple PI sources with different costs and informativeness, and allocating them based on step-level supervision needs.
More related work are discussed in Appendix~\ref{app:related}.

\section{Conclusion}

We presented \method, which constructs step-aware privileged information for multi-turn agent self-distillation.
By combining environment feedback on all steps with LLM-generated corrective hints on identified error steps, \method produces adaptive token-level supervision: weak confirmatory signals on routine steps and strong corrective signals where they matter most.
Across three benchmarks and three model scales, \method improves on both pure RL and self-distillation baselines.

\clearpage
\section*{Limitations}

\paragraph{Cost of the LLM analyzer.}
\method calls an external LLM on failed trajectories to identify error steps and write corrective hints. This adds an inference cost to training that GRPO does not pay, and it makes the method's behaviour depend on the analyzer's quality. Table~\ref{tab:analyzer} shows that the gain survives swapping the analyzer for three weaker models, but all four are large proprietary or frontier-scale systems; whether a small open model can play the role is untested. The cost is confined to training. Inference uses no privileged information, no LLM calls, and no environment-feedback injection.

\paragraph{Dependence on error step identifiability.}
The effectiveness of AHEAD's corrective hints relies on the LLM analyzer's ability to correctly identify which steps were critical errors. Our evaluation focuses on environments with relatively discrete, identifiable mistakes such as picking up the wrong object, navigating to an irrelevant location, or issuing an unproductive search query. In environments where errors are subtle, cumulative, or arise from omissions rather than overt wrong actions, the analyzer may fail to pinpoint the true decision points, reducing the quality of step-aware PI\@. How well the approach transfers to such settings remains an open question.

\paragraph{Scope of evaluation.}
Our experiments cover three benchmarks spanning embodied reasoning, web navigation, and search-augmented QA\@. While these represent diverse agent interaction patterns, they share relatively short trajectories with clear binary success/failure outcomes. Environments with longer horizons, continuous action spaces, partial observability, or soft reward signals may pose additional challenges for both error detection and PI construction.

\bibliography{references}
\bibliographystyle{iclr2027_conference}

\clearpage
\appendix
{\LARGE\bfseries Appendix\par}

\section{Related Work}
\label{app:related}

\paragraph{Reinforcement Learning for LLM Agents.}
Reinforcement learning is now widely used for post-training language model agents in interactive environments \citep{grpo,dong2025arpo,feng2025gigpo}. Agents trained with RL interact with environments over many steps, making sequential decisions in settings such as embodied reasoning \citep{shridhar2020alfworld}, web navigation \citep{yao2022webshop}, search-augmented QA \citep{jin2025searchr1}, and GUI automation \citep{lu2025uis1,lu2026uir1}. A central difficulty in these settings is credit assignment: outcome rewards indicate whether an episode succeeded but provide no information about which intermediate decisions were responsible. Our work addresses this by introducing step-aware dense supervision that complements the coarse trajectory-level signal.

\paragraph{Environment Feedback as Supervision.}
Standard agent RL discards environment observations during optimization, using them only as context for future actions. Recent work has recognized that these observations contain useful supervision signals. ECHO \citep{shrivastava2026echo} trains the policy to predict environment observations as an auxiliary task, learning an implicit world model at no additional rollout cost. EnvRL \citep{wang2026envrl} extends this with separate state prediction and inverse dynamics objectives. These methods treat environment feedback as prediction targets for representation learning. Our work takes a different perspective: we use environment feedback as privileged information for the teacher branch, enabling grounded token-level guidance. We further combine it with LLM-generated corrective hints on error steps, providing corrective direction that environment feedback alone cannot supply.

\paragraph{Privileged Information in Agent Training.}
Using privileged information during training while removing it at inference has roots in the learning-by-cheating paradigm \citep{chen2019cheating}. In LLM agent training, this idea appears as skill-conditioned learning, where natural-language skills are provided during training but removed at test time \citep{lu2026skill0,xia2026skillrl,shi2026skill1}. These approaches typically use a single type of privileged information applied uniformly across steps. Our work extends this line by combining multiple PI sources with different costs and informativeness, and allocating them based on step-level supervision needs.
\section{Dataset and Metric Details}
\label{app:datasets}

Table~\ref{tab:datasets} summarizes the datasets used in our experiments. Below we provide additional details on each benchmark and the corresponding evaluation metrics.

\begin{table}[!htp]
  \centering
  \small
  \begin{tabular}{lcc}
    \toprule
    \textbf{Benchmark} & \textbf{\#Train} & \textbf{\#Test} \\
    \midrule
    ALFWorld & 2,400 & 128 (Val-128) / 140 (Seen-140) / 134 (Unseen-134) \\
    \midrule
    WebShop & 2,400 & 128 \\
    \midrule
    NQ, TriviaQA, PopQA, HotpotQA, & 19,200 & 51,713 \\
    2WikiMultiHopQA, MuSiQue, Bamboogle & & \\
    \bottomrule
  \end{tabular}
  \caption{Datasets used in our experiments: ALFWorld (embodied reasoning), WebShop (web navigation), and Search-based QA. For Search-based QA we train only on NQ and HotpotQA; the remaining five datasets are held out and never seen during training.}
  \label{tab:datasets}
\end{table}

\paragraph{ALFWorld.}
ALFWorld \citep{shridhar2020alfworld} connects text-based interaction with the ALFRED household environment. The agent receives a natural-language goal and textual observations, and must issue a sequence of valid actions to complete the task. The benchmark covers six task categories: Pick, Look, Clean, Heat, Cool, and Pick2. We sample 2,400 training examples from the GiGPO split \citep{feng2025gigpo}. Evaluation uses three sets: \textbf{Val-128}, a fixed 128-task subset of the seen split that all baselines report on and that we use by default; \textbf{Seen-140}, the full seen split; and \textbf{Unseen-134}, the unseen split, which measures generalisation to room layouts not encountered during training. On each set we report the micro-averaged success rate, in which every task counts equally:
\begin{equation}
    \text{ALFWorld-Avg} = \frac{\sum_{c=1}^{6} n_c \, \text{SR}_c}{\sum_{c=1}^{6} n_c},
\end{equation}
where $n_c$ is the number of tasks in category $c$. Because the categories are not equally sized, this differs from the unweighted mean of the six per-category rates.

\paragraph{WebShop.}
WebShop \citep{yao2022webshop} is a simulated e-commerce environment where an agent searches for products, navigates product pages, selects attributes, and makes a purchase to satisfy a natural-language request. We use 2,400 training examples and evaluate on 128 fixed tasks following \citet{feng2025gigpo}. The environment returns two metrics: a normalized task-completion score that gives partial credit for matching requested attributes, and a binary success indicator for exact task completion. We report Score (the mean normalized score multiplied by 100) and Succ.\ (the percentage of exactly successful tasks).

\paragraph{Search-based QA.}
Following Search-R1 \citep{jin2025searchr1}, the agent interacts with a search engine to retrieve relevant documents before producing a final answer. The evaluation spans seven datasets: three single-hop (NQ \citep{kwiatkowski2019nq}, TriviaQA \citep{joshi2017triviaqa}, PopQA \citep{mallen2023popqa}) and four multi-hop (HotpotQA \citep{yang2018hotpotqa}, 2WikiMultiHopQA \citep{ho2020twowiki}, MuSiQue \citep{trivedi2022musique}, Bamboogle \citep{press2023bamboogle}). We train on 19,200 examples drawn from NQ and HotpotQA; the remaining five datasets serve as out-of-domain evaluation. We compute answer accuracy on each dataset and report the unweighted macro-average:
\begin{equation}
    \text{Search-Avg} = \frac{1}{7} \sum_{d=1}^{7} \text{Acc}_d.
\end{equation}
Unlike ALFWorld, where the six categories partition a single task suite and are therefore pooled, the seven QA datasets are independently constructed benchmarks with test sets of very different sizes; averaging them without weights keeps a large dataset such as TriviaQA from dominating the score.

\section{Baseline Descriptions}
\label{app:baselines}

Following \citet{sdar}, we adopt the same set of baselines for a fair comparison. All methods share the same backbone model, environment configuration, rollout budget, and evaluation setup, as well as the number of optimization steps, group size, and learning rate schedule. An asterisk (*) marks methods that retain the retrieved skill in the prompt at validation/test time.

\subsection{Prompting-Only Methods}

\paragraph{Vanilla.} The base instruction-tuned model is evaluated as-is, with no post-training applied. It receives the environment prompt and interaction history as its only input.

\paragraph{Skill-Prompt*.} No post-training is applied. Instead, a retrieved task-relevant skill is added to the prompt at validation/test time. Any performance gain therefore comes from the model's ability to leverage the skill in context.

\subsection{Outcome-Based Reinforcement Learning}

\paragraph{GRPO.} GRPO \citep{grpo} is a critic-free policy-gradient method. It generates multiple trajectories per task and computes advantages by normalizing their outcome rewards relative to the group. All tokens in a trajectory share the same sequence-level advantage. The policy is optimized with a clipped importance-ratio objective.

\paragraph{Skill-GRPO.} This method trains with the same GRPO objective but additionally includes a task-relevant skill in the prompt during rollouts. At validation/test time the skill is removed. This evaluates whether the policy has learned from skill-guided exploration well enough to perform without the skill.

\paragraph{Skill-GRPO*.} Training is the same as Skill-GRPO. The difference is that the skill remains in the prompt during validation/test time, so the model has consistent access to the skill in both training and evaluation.

\subsection{Self-Distillation and Hybrid Methods}

\paragraph{OPSD.} OPSD \citep{opsd} is a self-distillation method where the student and teacher are initialized from the same model. The two branches differ in their conditioning: the student sees only the task prompt, while the teacher has access to privileged information (e.g., a ground-truth solution). The teacher provides token-level distributional targets along the student's own trajectory, and only the student receives gradient updates. No privileged context is used at inference time.

\paragraph{Skill-SD.} Skill-SD \citep{wang2026skillsd} applies self-distillation to multi-turn agent settings. It first distills successful trajectories into natural-language skills that describe effective strategies and common failure modes. During training, these skills serve as privileged context for the teacher, while the student operates with the standard task prompt only. Since the skills are not available at test time, the student must learn to reproduce the teacher's behavior from its parameters alone.

\paragraph{GRPO+OPSD.} This baseline adds the OPSD distillation loss as an auxiliary term to the GRPO objective. The outcome-based component operates at the trajectory level, while the distillation component provides token-level supervision from a frozen reference. This tests whether a straightforward combination of the two signals is effective.

\paragraph{RLSD.} RLSD \citep{rlsd} modifies GRPO by using a privileged self-teacher to assign token-level importance weights. The log-probability gap between teacher and student at each token is mapped to a bounded scalar that adjusts the magnitude of that token's gradient update, while the sign still follows the outcome-level advantage. The teacher's contribution is scheduled to decay over training, so that later stages reduce to standard GRPO.

\paragraph{SDAR.} SDAR \citep{sdar} augments GRPO with a gated distillation loss as an auxiliary objective. A teacher conditioned on privileged context (e.g., a retrieved skill) evaluates the student's on-policy tokens and produces token-level signals. These signals pass through a bounded gate that upweights reliable positive guidance and downweights noisy negative ones before entering the loss. The GRPO advantage is not modified; the gating operates solely on the distillation term.

\section{Algorithm}
\label{app:algorithm}

The full procedure of AHEAD is presented in Algorithm~\ref{alg:ahead}. Compared to standard GRPO (which corresponds to removing lines~\ref{line:filter_start}--\ref{line:filter_end} and setting $\tilde{A}_{t,\ell} = A^{\mathrm{ep}}$ everywhere), AHEAD adds three components: (1)~an LLM analyzer call on failed trajectories, (2)~a PI-augmented forward pass to compute $\delta_{t,\ell}$, and (3)~a bounded reweight of the advantage. All three are confined to training; at inference time the policy acts from $h_t$ alone.

\begin{algorithm}[t]
\caption{AHEAD}
\label{alg:ahead}
\footnotesize
\begin{algorithmic}[1]
\Require Policy $\pi_\theta$, task set $\mathcal{S}$, group size $N$, clip bound $\epsilon$, reweight bound $\varepsilon$, initial mixing coefficient $\lambda_0$, decay horizon $D$, LLM analyzer $\mathcal{A}$
\For{each training iteration $k$}
    \State $\lambda_k \gets \max(0,\; \lambda_0 \cdot (1 - k / D))$ \Comment{\textcolor{algcomment}{\textit{Linear decay}}}
    \State Sample a batch of tasks $\{q\}$ from $\mathcal{S}$
    \For{each task $q$}
        \State \textcolor{algcomment}{\textit{// Stage 0: On-policy rollout}}
        \State Sample $N$ trajectories $\{\tau^{(1)}, \ldots, \tau^{(N)}\} \sim \pi_\theta(\cdot \mid q)$
        \State Compute $A^{\mathrm{ep}}_i = (R(\tau^{(i)}) - \mu_q) / \sigma_q$ \Comment{\textcolor{algcomment}{\textit{Group-relative advantage}}}
        \For{$i = 1, \ldots, N$}
            \If{$R(\tau^{(i)})$ indicates failure} \label{line:filter_start}
                \State \textcolor{algcomment}{\textit{// Stage 1: Error Step Detection}}
                \State $\mathcal{E}_\tau,\; \{\Phi^{\mathrm{llm}}_t\}_{t \in \mathcal{E}_\tau} \gets \mathcal{A}(\tau^{(i)})$ \Comment{\textcolor{algcomment}{\textit{LLM analyzer}}}
                \State \textcolor{algcomment}{\textit{// Stage 2: Step-Aware PI Construction}}
                \For{each step $t$ in $\tau^{(i)}$}
                    \State $\Phi^{\mathrm{env}}_t \gets o_{t+1}$ \Comment{\textcolor{algcomment}{\textit{Environment feedback}}}
                    \If{$t \in \mathcal{E}_\tau$}
                        \State $\tilde{h}_t \gets H(h_t,\, \Phi^{\mathrm{env}}_t,\, \Phi^{\mathrm{llm}}_t)$ \Comment{\textcolor{algcomment}{\textit{Env + Hint}}}
                    \Else
                        \State $\tilde{h}_t \gets H(h_t,\, \Phi^{\mathrm{env}}_t)$ \Comment{\textcolor{algcomment}{\textit{Env only}}}
                    \EndIf
                \EndFor
                \State \textcolor{algcomment}{\textit{// Stage 3: Token-Level Self-Distillation}}
                \For{each step $t$, token $\ell$}
                    \State $\delta_{t,\ell} \gets \log \pi_{\theta_{\mathrm{old}}}(y_{t,\ell} \mid \tilde{h}_t, y_{t,<\ell}) - \log \pi_{\theta_{\mathrm{old}}}(y_{t,\ell} \mid h_t, y_{t,<\ell})$
                    \State $w_{t,\ell} \gets \mathrm{clip}\!\big(\exp\!\big(\mathrm{sgn}(A^{\mathrm{ep}}) \cdot \mathrm{sg}(\delta_{t,\ell})\big),\; 1{-}\varepsilon,\; 1{+}\varepsilon\big)$
                    \State $\tilde{A}_{t,\ell} \gets A^{\mathrm{ep}} \cdot \big[(1 - \lambda_k) + \lambda_k \cdot w_{t,\ell}\big]$
                \EndFor
            \Else
                \State $\tilde{A}_{t,\ell} \gets A^{\mathrm{ep}}$ for all $t, \ell$ \Comment{\textcolor{algcomment}{\textit{Vanilla advantage}}} \label{line:filter_end}
            \EndIf
        \EndFor
        \State \textcolor{algcomment}{\textit{// Policy update}}
        \State Update $\theta$ by minimizing $\mathcal{L}(\theta) = -\mathbb{E}\!\left[\min\!\left(\rho_{t,\ell}\, \tilde{A}_{t,\ell},\; \hat{\rho}_{t,\ell}\, \tilde{A}_{t,\ell}\right)\right]$
    \EndFor
\EndFor
\end{algorithmic}
\end{algorithm}

\section{Additional Results}\label{app:results}

The main text reports the analysis of Sec.~\ref{sec:analysis} on Qwen2.5-7B\@. This appendix gives the per-category numbers behind it and repeats each experiment on Qwen2.5-3B and Qwen3-1.7B\@. The trends observed at 7B carry over to 1.7B, where the gains are larger. At 3B the average gains are smaller, because GRPO already scores $81.4$ on Seen-140, and the per-category changes are mixed.

\subsection{Per-Category Breakdown}

\begin{figure}[h]
  \centering
  \begin{minipage}[t]{0.49\textwidth}
    \centering
    \includegraphics[width=\linewidth]{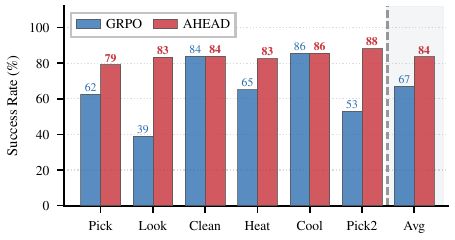}
    \caption{Per-task success rate on the ALFWorld \textbf{Unseen-134} split (Qwen2.5-7B-Instruct). \method raises the average from $67.2$ to $83.6$. The gains concentrate on the categories where GRPO is weakest, Look ($38.9 \rightarrow 83.3$) and Pick2 ($52.9 \rightarrow 88.2$), while categories GRPO already solves (Clean, Cool) are left unchanged.}
    \label{fig:bytask}
  \end{minipage}\hfill
  \begin{minipage}[t]{0.49\textwidth}
    \centering
    \includegraphics[width=\linewidth]{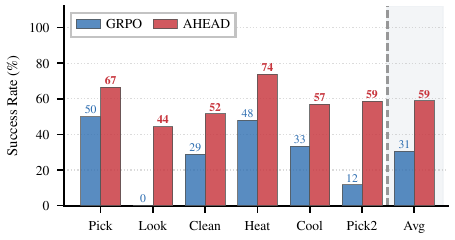}
    \caption{Per-task success rate on the ALFWorld \textbf{Unseen-134} split (Qwen3-1.7B-Instruct). Companion to Figure~\ref{fig:bytask}, at the smaller scale.}
    \label{fig:bytask_1p7b}
  \end{minipage}
\end{figure}

Table~\ref{tab:bytask_seen} and Table~\ref{tab:bytask} report per-category success rates on Seen-140 and Unseen-134. Figure~\ref{fig:bytask} plots the 7B rows of Table~\ref{tab:bytask} and Figure~\ref{fig:bytask_1p7b} the 1.7B rows. Unlike Table~\ref{tab:main_results}, whose baseline numbers are taken from \citet{sdar} on Val-128, the GRPO rows here come from our own reproduced GRPO checkpoints, since \citet{sdar} does not report per-category or full-split results. GRPO and \method are therefore trained and evaluated in the same codebase, so the two are directly comparable.

The improvement is not uniform across task categories. On Unseen-134 at 7B, \method leaves Clean and Cool untouched, categories that GRPO already solves at $83.9\%$ and $85.7\%$, while lifting Look by $44.4$ points and Pick2 by $35.3$ points. These two categories require the longest action sequences and are the ones where a single misstep most often derails the episode, which is where step-aware corrective hints are most valuable. The same ordering holds at 1.7B: the largest gains fall on Pick2 ($+47.1$) and Look ($+44.4$, from a GRPO baseline that solves none of the $18$ tasks).

The gain does not shrink on the unseen split. On Qwen2.5-7B, \method improves the average by $+16.4$ points on Seen-140 ($93.6$ vs.\ $77.1$) and by exactly $+16.4$ points on Unseen-134 ($83.6$ vs.\ $67.2$). On Qwen3-1.7B the gains are $+24.3$ ($66.4$ vs.\ $42.1$) and $+28.4$ ($59.0$ vs.\ $30.6$), larger on the unseen split than on the seen one. Qwen2.5-3B shows the same direction more sharply: the average gain is only $+1.5$ on Seen-140 ($82.9$ vs.\ $81.4$) but $+10.5$ on Unseen-134 ($82.1$ vs.\ $71.6$). If \method were simply overfitting to the training room layouts, its advantage would shrink on unseen rooms. It does not. Since the corrective hints are used only during training and removed at inference time, the improvement comes from what the policy has learned, not from access to privileged information.

\begin{table}[t]
  \begin{minipage}[t]{0.49\textwidth}
    \centering
    \small
    \setlength{\tabcolsep}{4pt}
    \resizebox{\linewidth}{!}{%
    \begin{tabular}{lccccccc}
      \toprule
      \textbf{Method} & \textbf{Pick} & \textbf{Look} & \textbf{Clean} & \textbf{Heat} & \textbf{Cool} & \textbf{Pick2} & \textbf{Avg.} \\
      \midrule
      \multicolumn{8}{l}{\emph{Qwen2.5-3B-Instruct}} \\
      GRPO & 91.4 & \textbf{76.9} & 81.5 & 62.5 & \textbf{88.0} & 75.0 & 81.4 \\
      \rowcolor{oursrow} \method & \textbf{97.1} & 53.8 & \textbf{88.9} & \textbf{87.5} & 72.0 & \textbf{79.2} & \textbf{82.9} \\
      \rowcolor{deltarow} $\Delta$ & +5.7 & $-23.1$ & +7.4 & +25.0 & $-16.0$ & +4.2 & +1.5 \\
      \midrule
      \multicolumn{8}{l}{\emph{Qwen2.5-7B-Instruct}} \\
      GRPO & 82.9 & 61.5 & 96.3 & 93.8 & \textbf{80.0} & 41.7 & 77.1 \\
      \rowcolor{oursrow} \method & \textbf{100.0} & \textbf{84.6} & \textbf{100.0} & \textbf{100.0} & \textbf{80.0} & \textbf{91.7} & \textbf{93.6} \\
      \rowcolor{deltarow} $\Delta$ & +17.1 & +23.1 & +3.7 & +6.3 & +0.0 & +50.0 & +16.4 \\
      \midrule
      \multicolumn{8}{l}{\emph{Qwen3-1.7B-Instruct}} \\
      GRPO & 68.6 & 15.4 & 55.6 & 50.0 & 20.0 & 20.8 & 42.1 \\
      \rowcolor{oursrow} \method & \textbf{82.9} & \textbf{53.8} & \textbf{70.4} & \textbf{68.8} & \textbf{68.0} & \textbf{41.7} & \textbf{66.4} \\
      \rowcolor{deltarow} $\Delta$ & +14.3 & +38.5 & +14.8 & +18.8 & +48.0 & +20.8 & +24.3 \\
      \bottomrule
    \end{tabular}%
    }
    \caption{Success rate (\%) per task category on the ALFWorld \textbf{Seen-140} split. $\Delta$ is the absolute gain of \method over GRPO in percentage points, and \textbf{Avg.}\ is the micro-average over all 140 tasks.}
    \label{tab:bytask_seen}
  \end{minipage}\hfill
  \begin{minipage}[t]{0.49\textwidth}
    \centering
    \small
    \setlength{\tabcolsep}{4pt}
    \resizebox{\linewidth}{!}{%
    \begin{tabular}{lccccccc}
      \toprule
      \textbf{Method} & \textbf{Pick} & \textbf{Look} & \textbf{Clean} & \textbf{Heat} & \textbf{Cool} & \textbf{Pick2} & \textbf{Avg.} \\
      \midrule
      \multicolumn{8}{l}{\emph{Qwen2.5-3B-Instruct}} \\
      GRPO & 70.8 & \textbf{88.9} & \textbf{77.4} & 65.2 & \textbf{81.0} & 41.2 & 71.6 \\
      \rowcolor{oursrow} \method & \textbf{83.3} & 77.8 & \textbf{77.4} & \textbf{87.0} & 76.2 & \textbf{94.1} & \textbf{82.1} \\
      \rowcolor{deltarow} $\Delta$ & +12.5 & $-11.1$ & +0.0 & +21.8 & $-4.8$ & +52.9 & +10.5 \\
      \midrule
      \multicolumn{8}{l}{\emph{Qwen2.5-7B-Instruct}} \\
      GRPO & 62.5 & 38.9 & \textbf{83.9} & 65.2 & \textbf{85.7} & 52.9 & 67.2 \\
      \rowcolor{oursrow} \method & \textbf{79.2} & \textbf{83.3} & \textbf{83.9} & \textbf{82.6} & \textbf{85.7} & \textbf{88.2} & \textbf{83.6} \\
      \rowcolor{deltarow} $\Delta$ & +16.7 & +44.4 & +0.0 & +17.4 & +0.0 & +35.3 & +16.4 \\
      \midrule
      \multicolumn{8}{l}{\emph{Qwen3-1.7B-Instruct}} \\
      GRPO & 50.0 & 0.0 & 29.0 & 47.8 & 33.3 & 11.8 & 30.6 \\
      \rowcolor{oursrow} \method & \textbf{66.7} & \textbf{44.4} & \textbf{51.6} & \textbf{73.9} & \textbf{57.1} & \textbf{58.8} & \textbf{59.0} \\
      \rowcolor{deltarow} $\Delta$ & +16.7 & +44.4 & +22.6 & +26.1 & +23.8 & +47.1 & +28.4 \\
      \bottomrule
    \end{tabular}%
    }
    \caption{Success rate (\%) per task category on the ALFWorld \textbf{Unseen-134} split. $\Delta$ is the absolute gain of \method over GRPO in percentage points, and \textbf{Avg.}\ is the micro-average over all 134 tasks.}
    \label{tab:bytask}
  \end{minipage}
\end{table}

\subsection{Training Dynamics}

\begin{figure}[h]
  \centering
  \includegraphics[width=\textwidth]{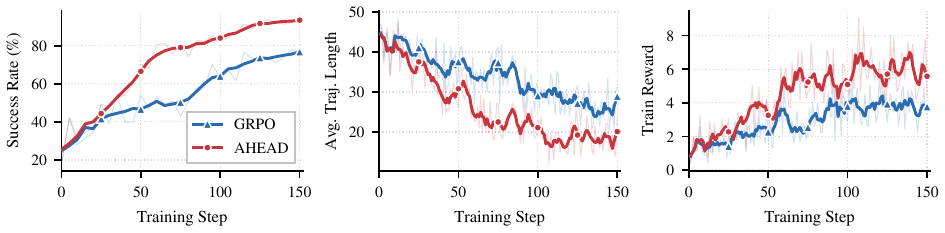}
  \caption{Training dynamics on ALFWorld (7B); the success-rate panel is measured on the seen split. AHEAD attains a higher success rate and train reward throughout training, while driving average trajectory length down faster. The agent solves more tasks and does so in fewer steps.}
  \label{fig:dynamics}
\end{figure}

\begin{figure}[h]
  \centering
  \includegraphics[width=\textwidth]{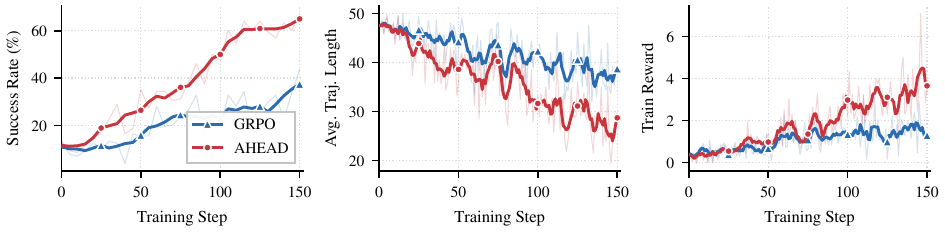}
  \caption{Training dynamics on ALFWorld (Qwen3-1.7B-Instruct); the success-rate panel is measured on the seen split. Companion to Figure~\ref{fig:dynamics}.}
  \label{fig:dynamics_1p7b}
\end{figure}

Figure~\ref{fig:dynamics} tracks training at 7B and Figure~\ref{fig:dynamics_1p7b} at 1.7B\@.

\method not only achieves higher final scores but also produces different training behavior. Success rate and train reward are consistently higher throughout optimization, with \method reaching a final train reward of 4.40 compared to 3.47 for GRPO. The trajectory length is particularly informative: both methods start at an average of 46.8 steps, but by the end of training \method has driven it down to 23.8 while GRPO only reaches 31.2. The agent thus solves more tasks and does so in fewer steps. A policy that merely explored more aggressively would improve reward at the cost of longer episodes; \method improves reward \emph{and} shortens trajectories, indicating that the corrective hints help the agent avoid the wasted actions that typically follow an error step. This pattern is consistent with the mechanism of step-aware PI\@: by providing corrective direction at error steps during training, the policy learns to avoid the common failure mode of repeating or escalating an incorrect action, which in standard GRPO often extends trajectories without progress.

The same picture holds at 1.7B\@. Both methods start from an average episode length of $47.8$ steps; by step $150$ \method has driven it down to $29.8$ while GRPO stalls at $40.2$, and it does so at a higher train reward ($3.62$ vs.\ $1.36$). Success rises as trajectory length falls, the same pattern observed at 7B but a sharper one: the $10.4$-step separation between the two methods at 1.7B exceeds the $7.4$ steps at 7B.

\subsection{Where the Reweighting Signal Concentrates}
\label{app:signal}

\paragraph{Setup.}
AHEAD assumes the self-distillation gap $\delta_{t,\ell}$ is larger on error
steps than on routine steps, so that reweighting the advantage puts more
credit on the tokens that caused the failure. We check this directly. We freeze
a mid-training checkpoint (step~25) on ALFWorld and collect $\delta_{t,\ell}$ on
the reweighted steps $\mathcal{M}_\tau$ (Eq.~\ref{eq:filter}): steps of failed
trajectories with non-zero advantage, i.e.\ the steps that actually affect the
loss. A step is an \emph{error step} if it is in the LLM-identified set
$\mathcal{E}_\tau$, and \emph{routine} otherwise. This gives $2{,}525$ error
steps and $26{,}877$ routine steps.

\begin{figure}[t]
  \centering
  \begin{subfigure}{0.49\textwidth}
    \centering
    \includegraphics[width=\linewidth]{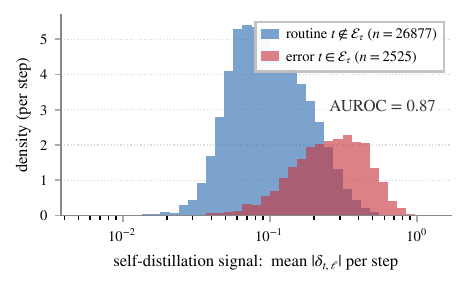}
    \caption{Self-distillation signal $|\delta_{t,\ell}|$.}
    \label{fig:dist-signal}
  \end{subfigure}\hfill
  \begin{subfigure}{0.49\textwidth}
    \centering
    \includegraphics[width=\linewidth]{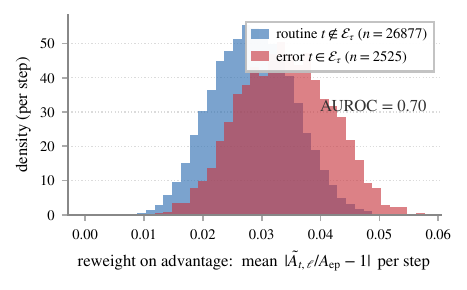}
    \caption{Reweight applied to the advantage.}
    \label{fig:dist-reweight}
  \end{subfigure}
  \caption{\textbf{The reweighting signal targets error steps.}
  Per-step distributions on the reweighted steps $\mathcal{M}_\tau$ (ALFWorld,
  frozen step-25 checkpoint). \textbf{(a)} The gap $|\delta_{t,\ell}|$ is larger
  on error steps $t\!\in\!\mathcal{E}_\tau$ than on routine steps (AUROC $0.87$).
  \textbf{(b)} After the clip and mixing $\lambda_k$, the reweight that actually
  scales the advantage, $|\tilde{A}_{t,\ell}/A_{\mathrm{ep}}\!-\!1|$, still
  favors error steps but with a smaller gap (AUROC $0.70$), leaving the reward in
  control of the update direction.}
  \label{fig:distributions}
\end{figure}

\paragraph{The signal is larger on error steps.}
Figure~\ref{fig:dist-signal} shows the per-step $|\delta_{t,\ell}|$ for both
groups. Error steps have a larger gap (mean $0.37$ vs.\ $0.17$), and the two
distributions separate clearly: ranking steps by $|\delta_{t,\ell}|$ recovers
the LLM error labels at AUROC $0.87$. So the richer PI on error steps
(Eq.~\ref{eq:pi_injection}) makes the signal stronger there on its own.

\paragraph{The reweight in the loss is smaller but still targets error steps.}
We do not apply the raw gap. The clip ($\epsilon\!=\!0.2$) and the mixing
coefficient $\lambda_k$ bound and scale it
(Eq.~\ref{eq:weight}--\ref{eq:reweight}).
Figure~\ref{fig:dist-reweight} shows what the advantage is actually multiplied
by, $|\tilde{A}_{t,\ell}/A_{\mathrm{ep}}\!-\!1|$. Error steps are still favored,
but the gap shrinks (AUROC $0.70$), leaving the reward in control of the update
direction.

\begin{figure}[t]
  \centering
  \includegraphics[width=\textwidth]{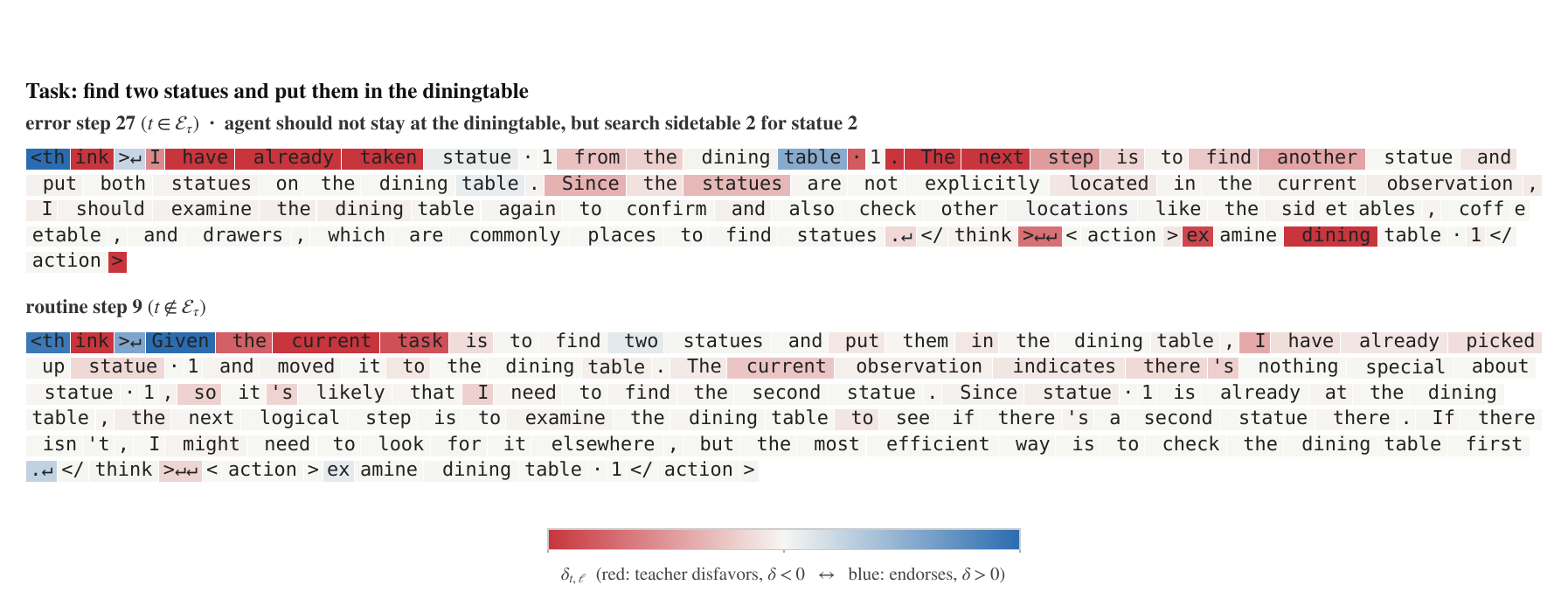}\\[3pt]
  \includegraphics[width=\textwidth]{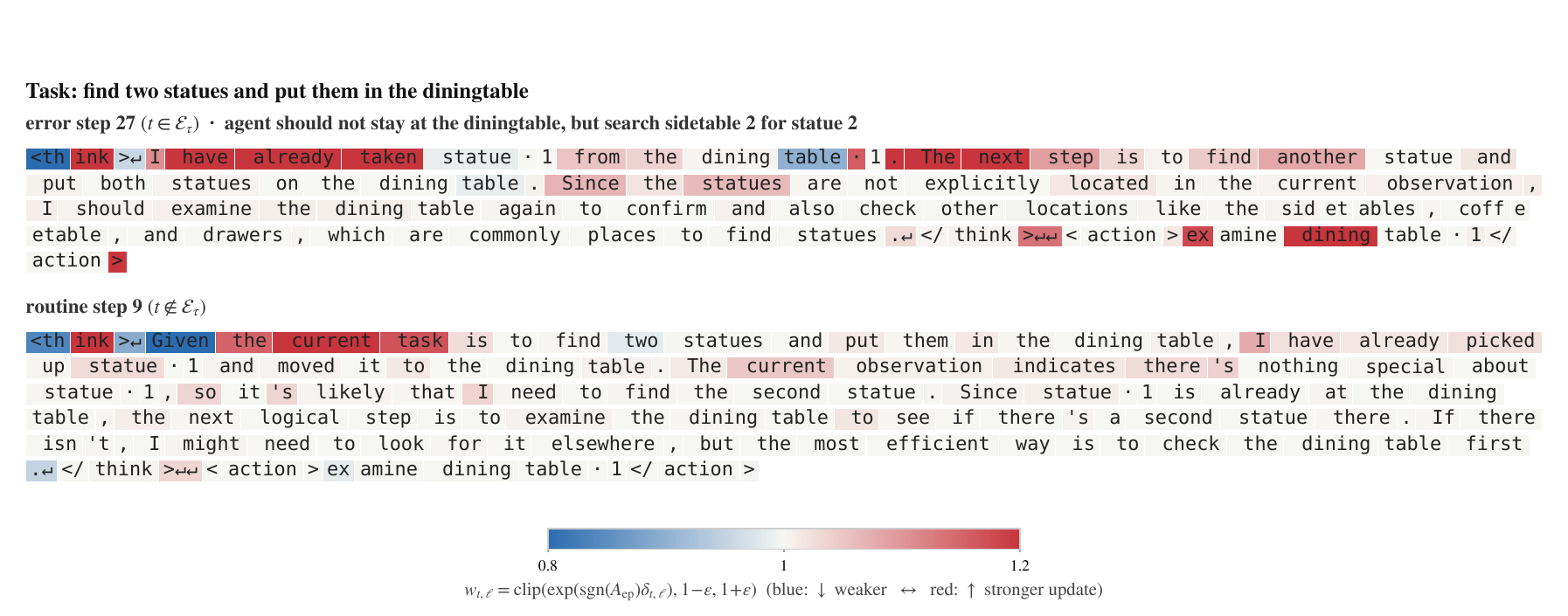}
  \caption{\textbf{Token-level credit on one failed trajectory} (ALFWorld).
  \textbf{Top:} signed gap $\delta_{t,\ell}$; red marks tokens the teacher
  disfavors ($\delta\!<\!0$). At the error step ($t\!\in\!\mathcal{E}_\tau$) the
  tokens for the wrong decision have the largest $|\delta|$; the routine step
  ($t\!\notin\!\mathcal{E}_\tau$) stays near zero. \textbf{Bottom:} the weight
  $w_{t,\ell}=\mathrm{clip}(\exp(\mathrm{sgn}(A_{\mathrm{ep}})\delta_{t,\ell}),
  1\!-\!\epsilon,1\!+\!\epsilon)$. Since $A_{\mathrm{ep}}\!<\!0$ on failed
  trajectories, these tokens get the largest amplification ($w\!>\!1$).}
  \label{fig:case-study}
\end{figure}

\paragraph{Token-level view.}
Figure~\ref{fig:case-study} shows one failed trajectory (task: \emph{find two
statues and put them in the diningtable}). At the error step
($t\!\in\!\mathcal{E}_\tau$), the tokens for the wrong decision---re-examining
the diningtable instead of searching \texttt{sidetable 2}---have the largest
$|\delta_{t,\ell}|$. Since $A_{\mathrm{ep}}\!<\!0$ on failed trajectories, these
are exactly the tokens amplified by $w_{t,\ell}\!>\!1$ (Eq.~\ref{eq:weight}). A
routine step in the same trajectory stays near uniform. This shows the
reweighting localizes credit through $\delta$ alone.

\clearpage

\section{Hyperparameters}
\label{app:hyper}

Table~\ref{tab:hyper} lists the full configuration. All backbones share these settings. Note that \method adds no KL regularisation term and no per-step coefficient: the only method-specific hyperparameters are the reweight bound $\varepsilon$ and the decay schedule of $\lambda_k$.

\begin{table}[t]
  \centering
  \small
  \begin{tabular}{ll}
    \toprule
    \textbf{Hyperparameter} & \textbf{Value} \\
    \midrule
    Training steps & 150 \\
    Training batch size & 16 for ALFWorld and WebShop; 128 for Search-based QA \\
    Rollout group size $N$ & 8 \\
    Learning rate & $1 \times 10^{-6}$ \\
    Hardware & 8 $\times$ H100 \\
    GRPO clipping range $\epsilon$ & 0.2 \\
    \midrule
    Reweight bound $\varepsilon$ & 0.2 \\
    Initial distillation coefficient $\lambda_0$ & 0.5 \\
    Decay horizon $D$ & 50 \\
    LLM analyzer & Claude Opus~4.7 \\
    \midrule
    Maximum prompt length & 2,048 for ALFWorld; 4,096 for WebShop and Search-based QA \\
    Maximum interaction steps & 50 for ALFWorld; 15 for WebShop; 4 for Search-based QA \\
    \bottomrule
  \end{tabular}
  \caption{Training configuration. The block in the middle lists the parameters introduced by \method (Sec.~\ref{sec:reweight}); everything else is inherited unchanged from the GRPO baseline, so the comparison in Table~\ref{tab:main_results} isolates the effect of step-aware distillation.}
  \label{tab:hyper}
\end{table}

\section{Prompts}
\label{app:prompts}

\subsection{Environment Interaction Prompts}

The ALFWorld and WebShop templates follow \citet{feng2025gigpo}, and the Search-based QA template follows \citet{jin2025searchr1}. Every method (\method and all baselines alike) rolls out with the same template in a given environment, so the comparison in Table~\ref{tab:main_results} is not confounded by prompt wording. Braces mark slots filled at run time by the environment. Note that the agent is shown only a recent window of the interaction history rather than the full trajectory, making the step-level history $h_t$ in Sec.~\ref{sec:pi} a bounded context.

\begin{figure}[t]
\begin{promptbox}{Prompt for ALFWorld}
You are an expert agent operating in the ALFRED Embodied Environment. Your task is to: \slot{task\_description}\\
Prior to this step, you have already taken \slot{step\_count} step(s). Below are the most recent \slot{history\_length} observations and the corresponding actions you took: \slot{action\_history}\\
You are now at step \slot{current\_step} and your current observation is: \slot{current\_observation}\\
Your admissible actions of the current situation are: [\slot{admissible\_actions}].\\[5pt]
Now it's your turn to take an action.\\
You should first reason step-by-step about the current situation. This reasoning process MUST be enclosed within \xtag{think} \xtag{/think} tags.\\
Once you've finished your reasoning, you should choose an admissible action for current step and present it within \xtag{action} \xtag{/action} tags.
\end{promptbox}
\caption{Prompt template for the ALFWorld environment.}
\label{fig:prompt_alfworld}
\end{figure}

\begin{figure}[t]
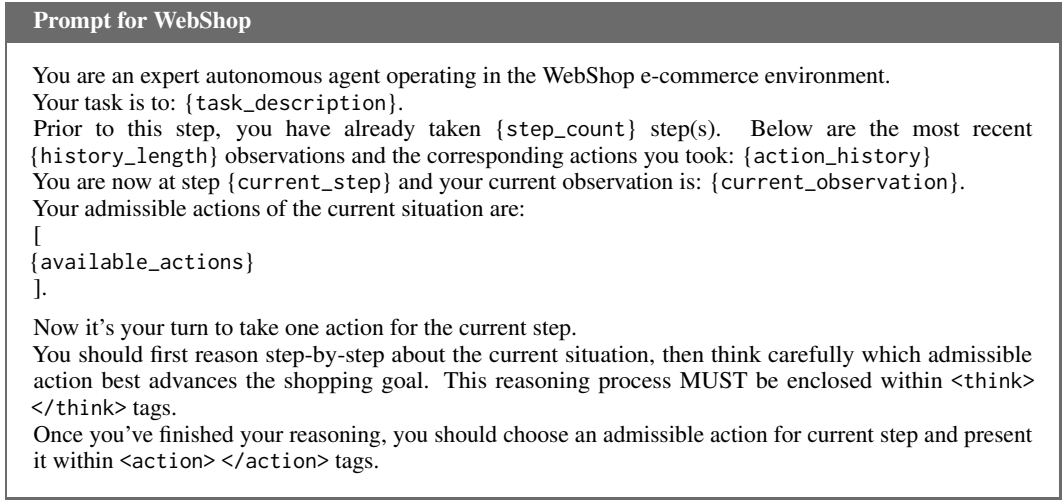

\begin{promptbox}{Prompt for WebShop}
You are an expert autonomous agent operating in the WebShop e-commerce environment.\\
Your task is to: \slot{task\_description}.\\
Prior to this step, you have already taken \slot{step\_count} step(s). Below are the most recent \slot{history\_length} observations and the corresponding actions you took: \slot{action\_history}\\
You are now at step \slot{current\_step} and your current observation is: \slot{current\_observation}.\\
Your admissible actions of the current situation are:\\
{}[\\
\slot{available\_actions}\\
].\\[5pt]
Now it's your turn to take one action for the current step.\\
You should first reason step-by-step about the current situation, then think carefully which admissible action best advances the shopping goal. This reasoning process MUST be enclosed within \xtag{think} \xtag{/think} tags.\\
Once you've finished your reasoning, you should choose an admissible action for current step and present it within \xtag{action} \xtag{/action} tags.
\end{promptbox}
\caption{Prompt template for the WebShop environment.}
\label{fig:prompt_webshop}
\end{figure}

\begin{figure}[t]
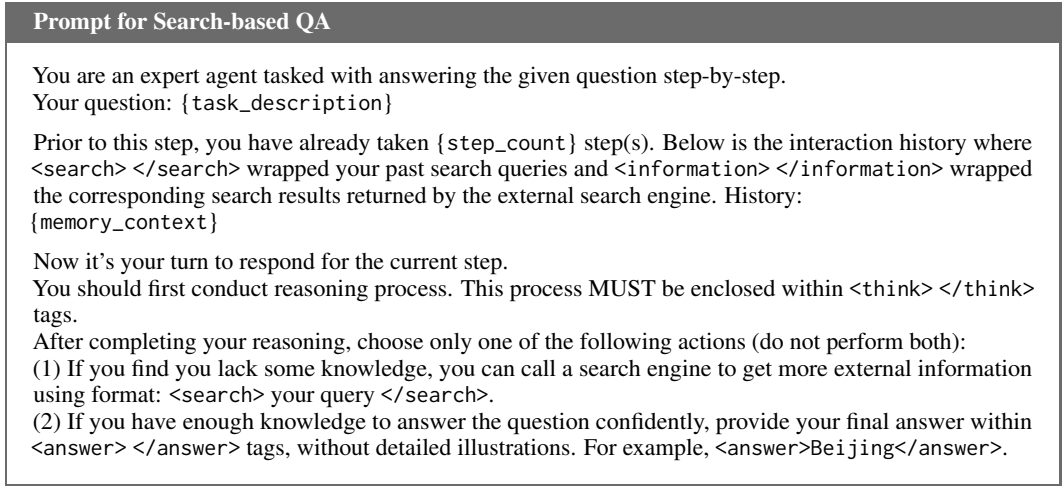

\begin{promptbox}{Prompt for Search-based QA}
You are an expert agent tasked with answering the given question step-by-step.\\
Your question: \slot{task\_description}\\[5pt]
Prior to this step, you have already taken \slot{step\_count} step(s). Below is the interaction history where \xtag{search} \xtag{/search} wrapped your past search queries and \xtag{information} \xtag{/information} wrapped the corresponding search results returned by the external search engine. History:\\
\slot{memory\_context}\\[5pt]
Now it's your turn to respond for the current step.\\
You should first conduct reasoning process. This process MUST be enclosed within \xtag{think} \xtag{/think} tags.\\
After completing your reasoning, choose only one of the following actions (do not perform both):\\
(1) If you find you lack some knowledge, you can call a search engine to get more external information using format: \xtag{search} your query \xtag{/search}.\\
(2) If you have enough knowledge to answer the question confidently, provide your final answer within \xtag{answer} \xtag{/answer} tags, without detailed illustrations. For example, \texttt{\textless answer\textgreater Beijing\textless /answer\textgreater}.
\end{promptbox}
\caption{Prompt template for the Search-based QA environment.}
\label{fig:prompt_search}
\end{figure}

\subsection{Error-Step Analyzer Prompt}

Figure~\ref{fig:prompt_analyzer} gives the prompt behind Stage~1 and Stage~2 of \method (Sec.~\ref{sec:pi}). A single call does both jobs: it selects the error steps $\mathcal{E}_\tau$ and, for each one, writes the corrective hint $\Phi^{\mathrm{llm}}_t$ that is injected into the teacher context. It is invoked only on failed trajectories.

Two constraints in the prompt matter for the method. First, the analyzer is asked for \emph{at most} \slot{max\_steps} key decision steps rather than a label for every step, which keeps $\mathcal{E}_\tau$ sparse so that most steps fall back to environment feedback alone. Second, the hint is required to be an imperative naming the correct object or query, and is explicitly forbidden from describing what went wrong. A hint that merely restated the failure would duplicate what the environment feedback already provides; the corrective prescription is what environment feedback cannot provide.

\begin{figure}[t]
\begin{promptbox}{Prompt for the Error-Step Analyzer}
\textbf{System:} You are an expert agent trajectory analyzer.\\[5pt]
\textbf{User:}\\
\texttt{\#\# Task}\\
\slot{task}\\[3pt]
\texttt{\#\# Failed Trajectory}\\
\slot{failed}\\
\slot{success\_block}\\[3pt]
\texttt{\#\# Instructions}\\
You are given a FAILED trajectory. Each step is labeled \texttt{[step N] action: ...\ \textbar{} env: ...}; N is that step's index. Indices are 0-based and contiguous (first step is \texttt{[step 0]}); valid indices are EXACTLY the N values printed above. \slot{noref\_note}\\
Read the ENTIRE trajectory (and the reference Successful Trajectory, if provided) and judge each step against the \texttt{\#\# Task} goal. Select AT MOST \slot{max\_steps} KEY decision steps: the steps where the choice of action most determined whether the task would succeed. Prefer the steps that need correcting; do NOT list every step, only the few that mattered most. \texttt{first\_fatal\_error} = the earliest step that made failure unavoidable.\\
Rules: each \texttt{step\_index} must be copied verbatim from a printed \texttt{[step N]} label (do NOT add/subtract 1).\\
Respond ONLY with JSON (no prose, no markdown fences):\\[3pt]
\texttt{\{"error\_steps": [\{"step\_index": \textless int\textgreater, "diagnosis": "\textless one sentence: why this step is a key decision\textgreater", "hint": "\textless provide one imperative, concise sentence telling the agent the correct action to take at this step; do NOT describe what went wrong or use the word 'agent'\textgreater"\}, ...], "first\_fatal\_error": \textless int\textgreater\}}
\end{promptbox}
\caption{Prompt template for the LLM error-step analyzer.}
\label{fig:prompt_analyzer}
\end{figure}

\end{document}